%% file: main.tex
\documentclass[letterpaper]{article}
\usepackage[preprint]{aaai2027}
\usepackage[hyphens]{url}
\usepackage{graphicx}
\usepackage{natbib}
\usepackage{caption}
\usepackage{algorithm}
\usepackage{algorithmic}
\usepackage{amsmath}
\usepackage{amssymb}
\usepackage{multirow}
\usepackage{booktabs}
\usepackage{newfloat}
\usepackage{listings}
\DeclareCaptionStyle{ruled}{labelfont=normalfont,labelsep=colon,strut=off}
\floatstyle{ruled}
\newfloat{listing}{tb}{lst}{}
\floatname{listing}{Listing}
\graphicspath{{./images/}}
\DeclareGraphicsExtensions{.pdf,.jpg,.png}

\title{LAD-COD: Language-Aligned Dense Perception \\ for Camouflaged Object Detection}

\author{
    Shangye Song\textsuperscript{\rm 1},
    Tianzhi Zhu\textsuperscript{\rm 2},
    Syed Ariff Syed Hesham\textsuperscript{\rm 3,4},
    Xin He\textsuperscript{\rm 5},
    Yun Liu\textsuperscript{\rm 2,6,7}\thanks{Corresponding author: Yun Liu (liuyun@nankai.edu.cn)}
}
\affiliations{
    \textsuperscript{\rm 1}School of Computer Science, The University of Auckland\\
    \textsuperscript{\rm 2}VCIP, College of Computer Science, Nankai University\\
    \textsuperscript{\rm 3}School of Electrical and Electronics Engineering, Nanyang Technological University, Singapore\\
    \textsuperscript{\rm 4}A*STAR Institute of Advanced Intelligence and Computing, Singapore\\
    \textsuperscript{\rm 5}School of Computer Science and Engineering, Tianjin University of Technology\\
    \textsuperscript{\rm 6}Academy for Advanced Interdisciplinary Studies, Nankai University\\
    \textsuperscript{\rm 7}Nankai International Advanced Research Institute, Shenzhen Futian
}

\begin{document}

\maketitle

\begin{abstract}
Camouflaged object detection (COD) aims to segment objects that exhibit high visual similarity to their surroundings, which reduces foreground-background discriminability and weakens boundary evidence across appearance, texture, and structure. Such limitations motivate the use of instruction-conditioned semantics as top-down guidance for identifying which weak visual cues are relevant to the target. Recent segmentation systems built on large multimodal models (LMMs) demonstrate this possibility through instruction-conditioned target embeddings that guide mask decoding. However, in this language-to-mask paradigm, the generated target embedding conditions mainly the mask decoder, leaving the dense visual features that must preserve low-contrast boundaries and fine local structure without explicit guidance. We propose Language-Aligned Dense perception for COD (\textbf{LAD-COD}), a framework that aligns top-down semantic target guidance with bottom-up hierarchical visual features. Instead of fully adapting a large generic image encoder, LAD-COD learns a trainable hierarchical visual branch that captures camouflage-sensitive texture, boundary, and contextual information. To align these features with the target embedding, LAD-COD applies Language-Aligned Dual Visual Fusion (\textbf{LADVF}), which extends the embedding beyond sparse prompting to query patch-level language-aligned features and to gate their residual integration with the hierarchical features. This design allows semantic information to guide localization while preserving the fine structural details needed for camouflage segmentation. Experiments on CAMO, COD10K, and NC4K show that LAD-COD obtains the best reported value in all 12 dataset-metric comparisons.
\end{abstract}

\input{sections/intro}
\input{sections/related}
\input{sections/method}
\input{sections/experiment}
\input{sections/conclusion}

\section*{Acknowledgements}
This work is supported by the National Natural Science Foundation of China (No. 62576176). The computational resources are supported by the Supercomputing Center of Nankai University (NKSC).

\bibliography{aaai2027}

\input{sections/supplement}

\end{document}

%% file: sections/intro.tex
\section{Introduction}

\begin{figure}[t]
    \centering
    \includegraphics[width=\columnwidth]{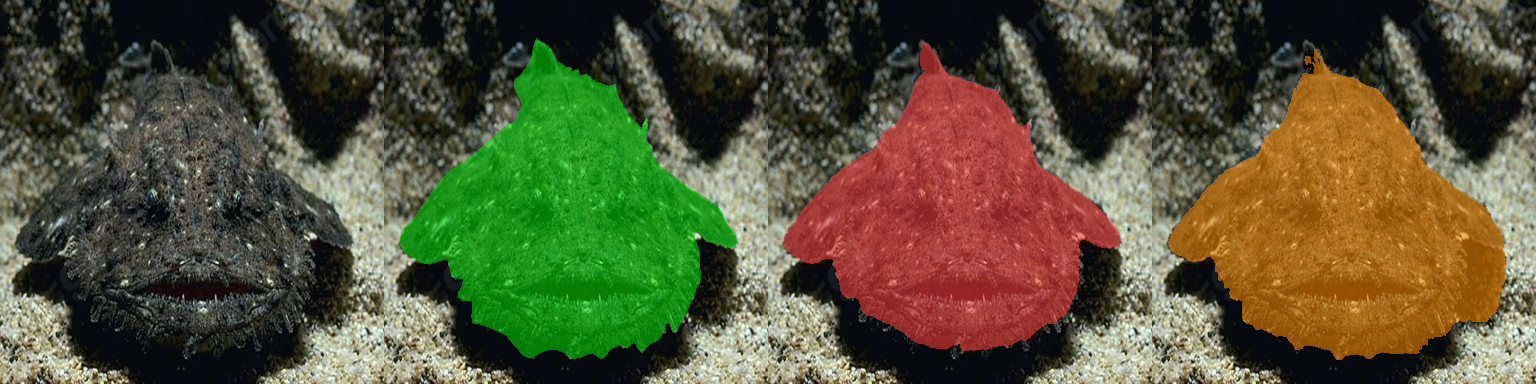}\\[0.02em]
    \includegraphics[width=\columnwidth]{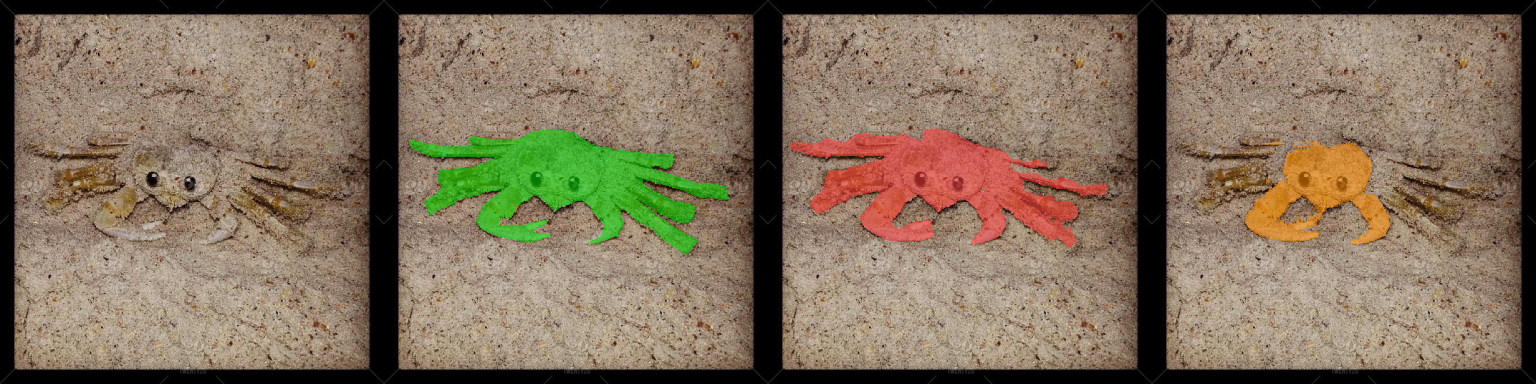}\\[0.02em]
    \includegraphics[width=\columnwidth]{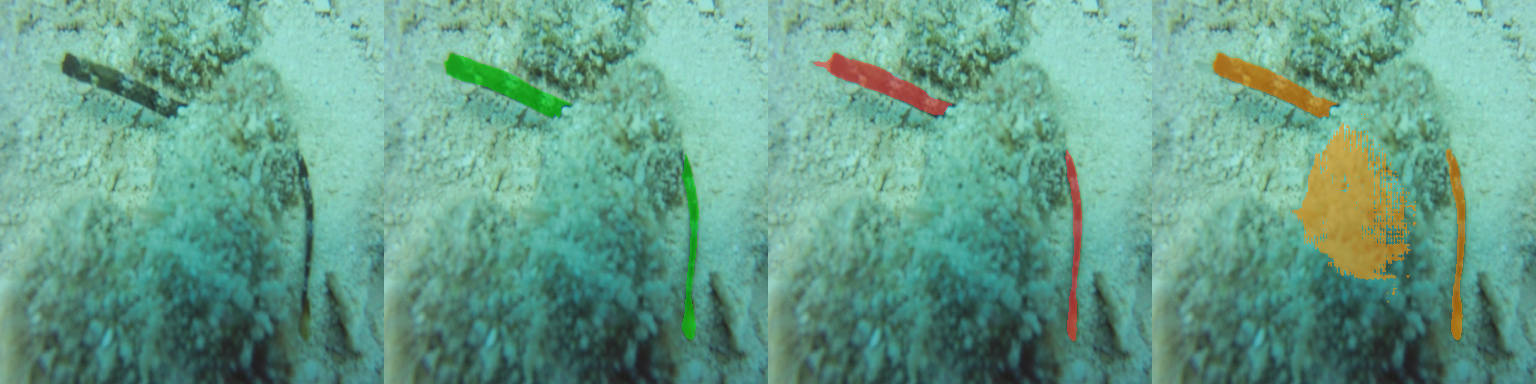}\\[0.02em]
    {\small
    \begin{minipage}[t]{0.25\columnwidth}\centering Input\end{minipage}%
    \begin{minipage}[t]{0.25\columnwidth}\centering GT\end{minipage}%
    \begin{minipage}[t]{0.25\columnwidth}\centering Ours\end{minipage}%
    \begin{minipage}[t]{0.25\columnwidth}\centering LISA\end{minipage}}
    \vspace{-5mm}
    \caption{Visual comparison on three COD samples. Both methods use the instruction ``Please segment the camouflaged object in this image.'' and are fine-tuned on the same COD data. LISA often misses weak boundaries, while LAD-COD (Ours) produces more complete masks.}
    \vspace{-3mm}
    \label{fig:intro}
\end{figure}

Camouflaged object detection (COD) aims to segment objects that blend into their surroundings. When foreground and background appearances are similar, the color, texture, and boundary contrasts that normally support localization become weak, and the remaining evidence is fragmented and easily confused with background variations \cite{Fan_2020_CVPR, pang2022zoom}. Existing COD methods address this weakness by strengthening bottom-up structural evidence. Multi-scale aggregation and contextual reasoning help discover concealed regions, while texture modeling, boundary supervision, and hierarchical features preserve the local variations for accurate delineation \cite{Fan_2020_CVPR, yan2021mirrornet, zhu2021inferring, Sun_2022, Ye_2025_ICCV}. Some other approaches further introduce structural cues such as depth \cite{wu2023source, wang2023depth}. Together these developments have substantially improved camouflage-sensitive perception without changing how the target is selected. Selection is still inferred from low-contrast appearance, and visually similar background structures can produce competing responses. The resulting representations preserve useful structure but provide no explicit instruction-conditioned target guidance.

Goal-directed human search offers a useful analogy: without a specified target, a camouflaged region rarely stands out, yet once the task is to find it, the image itself is unchanged while subtle patterns that would otherwise be overlooked become relevant. Instruction-conditioned semantics play the same role for a model, indicating which weak visual cues belong to the target. Segmentation built on large multimodal models (LMMs) supplies this top-down target guidance. Recent multimodal COD methods introduce language conditioning or automatically generated prompts into foundation-model-based segmentation \cite{ren2025multi, hu2024gensam}. Beyond COD, systems such as LISA~\cite{lai2024lisa}, GLaMM~\cite{rasheed2024glamm}, and PixelLM~\cite{ren2024pixellm} connect instruction following with dense prediction. In such LMM-to-SAM systems, the image-instruction pair produces a target embedding that serves as a sparse prompt for mask decoding. The embedding gives the decoder an explicit target-conditioned cue, but its role remains concentrated at the decoding stage, limiting its interaction with the dense visual representation for mask prediction.

In existing LMM-to-SAM systems, however, target guidance remains only loosely coupled with dense structural features. A sparse target prompt can steer mask decoding, but it does not explicitly organize the dense visual features the decoder receives. This distinction matters in COD, where accurate masks depend on weak boundaries and local structure that must be retained throughout visual encoding. Figure~\ref{fig:intro} provides a qualitative illustration. Under the same instruction and COD fine-tuning data, LISA often misses weak object boundaries. The central question is therefore how target guidance can shape camouflage-sensitive dense features without suppressing their structural detail.

We propose Language-Aligned Dense perception for COD (LAD-COD) to connect the two. Its instruction-conditioned target embedding provides top-down guidance, while a trainable hierarchical visual branch preserves bottom-up evidence across multiple scales. Rather than fully adapting a large generic image encoder, this branch learns features tailored to weak camouflage boundaries and local texture variations. The target embedding guides mask decoding as a sparse prompt and is further used by Language-Aligned Dual Visual Fusion (LADVF) to retrieve target-related information from patch-level language-aligned features. Gated residual fusion then injects the retrieved semantics into the hierarchical representation without replacing its structural content. Experiments on CAMO \cite{le2019anabranch}, COD10K \cite{Fan_2020_CVPR}, and NC4K \cite{lv2021ranknet} show that LAD-COD obtains the best reported value in all 12 dataset-metric comparisons, with the largest margins on objects embedded in closely matching texture, where top-down guidance contributes most. Controlled ablations further distinguish the contributions of target prompting, hierarchical dense features, and LADVF.

The main contributions are summarized as follows:
\begin{itemize}
    \item We introduce LAD-COD, a language-guided COD framework that combines instruction-conditioned target guidance with hierarchical dense visual representations.
    \item We develop LADVF, which extends the target embedding beyond mask prompting to retrieve target-related visual semantics and integrate them with hierarchical features through gated residual fusion.
    \item LAD-COD obtains the best reported value in all 12 dataset-metric comparisons across the three COD benchmarks, while controlled ablations distinguish the contributions of target prompting, four-stage hierarchical features, and LADVF.
\end{itemize}

\begin{figure*}[!t]
    \centering
    \includegraphics[width=\textwidth,height=0.39\textheight,keepaspectratio]{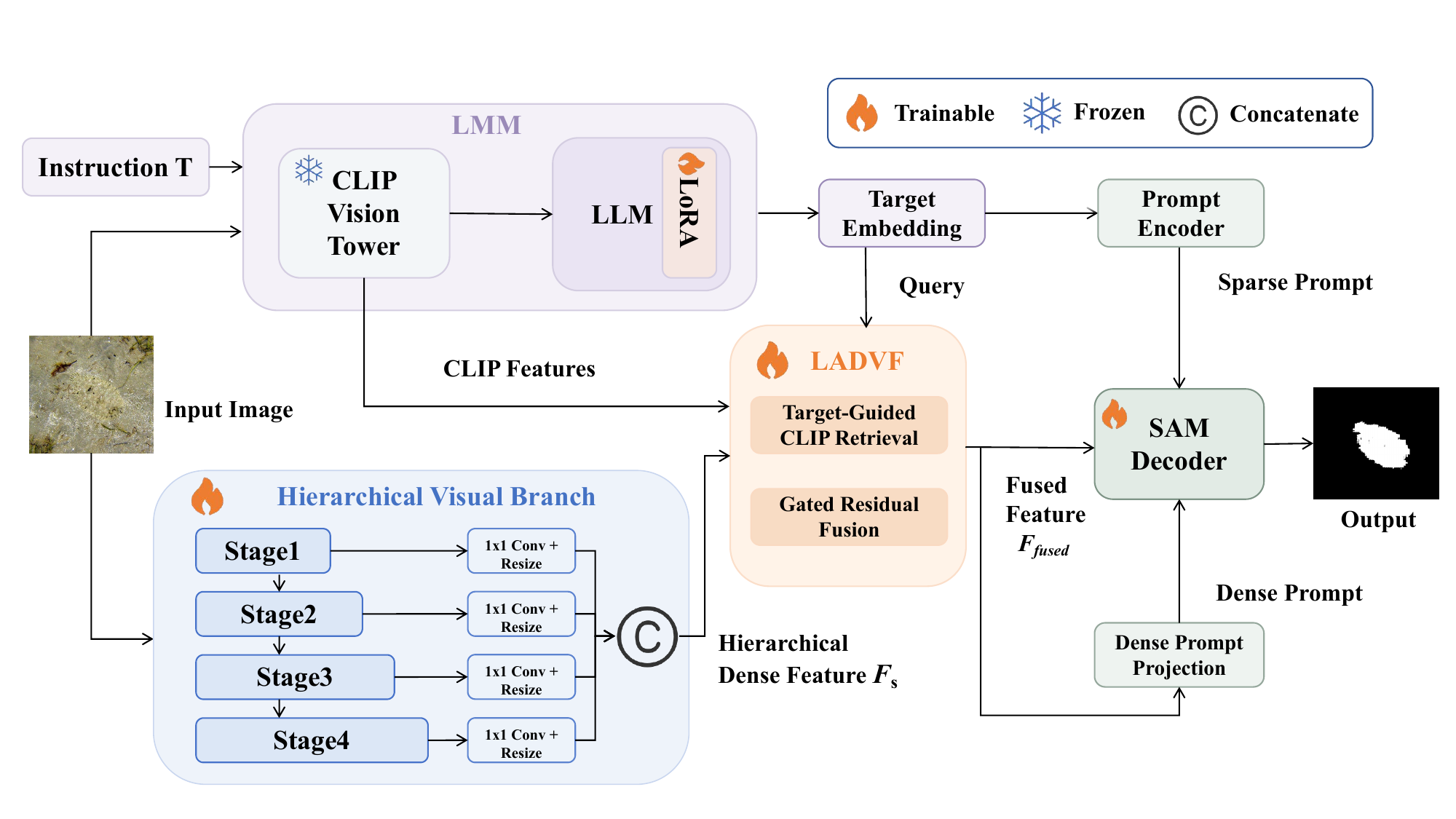}
    \caption{Overview of the proposed LAD-COD framework. The LMM produces an instruction-conditioned target embedding from the image-instruction pair, which guides sparse mask prompting and target-aware semantic retrieval in LADVF. A trainable hierarchical visual branch preserves camouflage-sensitive structural features, and gated residual fusion aligns the two representations before mask decoding.}
    \label{fig:pipeline}
\end{figure*}

%% file: sections/related.tex
\section{Related Work}

\subsection{Camouflaged Object Detection}
Camouflaged object detection (COD) aims to segment objects that blend into their surroundings and exhibit weak foreground-background contrast, making boundary localization difficult. SINet established a search-and-identification strategy, which SINet-V2 extended through neighbor connection and group-reversal attention \cite{Fan_2020_CVPR, fan2021concealed}. Subsequent methods strengthen localization through bio-inspired mirror-stream fusion \cite{yan2021mirrornet}, texture-aware interactive guidance \cite{zhu2021inferring}, context-aware cross-level fusion \cite{chen2022c2fnet}, mixed-scale reasoning \cite{pang2022zoom}, and multi-stage refinement \cite{zhang2022preynet, jia2022segmar}. Because contours provide important structural evidence under camouflage, recent methods exploit boundary supervision, boundary-and-texture enhancement, and edge-semantic collaboration to recover weak object structures \cite{Sun_2022, li2022findnet, Ye_2025_ICCV}. Depth provides an additional source of structural information when RGB appearance is ambiguous \cite{wu2023source, wang2023depth}. Other COD studies further explore uncertainty-guided reasoning \cite{yang2021ugtr}, gradient and texture learning \cite{ji2023dgnet}, joint salient-camouflaged object detection \cite{li2021jcod}, frequency-aware representations \cite{lin2023fbnet}, and distraction mining \cite{mei2021pfnet}.

With the rise of foundation segmentation models, COD research has begun to move beyond task-specific visual architectures. SAM-Adapter adapts SAM to COD through lightweight tuning \cite{chen2023samadapter}, while GenSAM and adaptive prompt generation methods derive visual prompts from task-level or semantic information to reduce manual prompting \cite{hu2024gensam, chen2025enhancing}. MM-SAM further incorporates language-guided multimodal information into a SAM-based system for camouflaged scene segmentation \cite{ren2025multi}. Together, these methods extend promptable segmentation and multimodal conditioning to COD, but leave unresolved how semantic conditioning should interact with visual representations that preserve weak boundaries and low-contrast structure. Vision-language segmentation examines this relationship more directly by connecting textual conditioning to dense visual representations for mask prediction.

\subsection{Vision-Language Segmentation}
Vision-language segmentation predicts pixel-level masks for objects or regions specified through natural language. A common route pairs language-aligned visual representations with a promptable mask decoder, roles typically filled by CLIP and SAM \cite{radford2021learning, kirillov2023segment}. Grounding DINO instead localizes text-matched regions, and Grounded SAM combines this grounding stage with SAM for open-world segmentation \cite{liu2023grounding, ren2024groundedsam}.

Recent segmentation methods based on large multimodal models (LMMs) connect instruction following more directly to mask prediction, building on the multimodal interaction enabled by visual instruction tuning in LLaVA \cite{liu2024visual}. LISA realizes this connection through a special \texttt{[SEG]} token whose hidden state conditions a mask decoder \cite{lai2024lisa}, while GLaMM and PixelLM extend LMM-based segmentation toward fine-grained grounding and pixel-level reasoning \cite{rasheed2024glamm, ren2024pixellm}. This language-to-mask design provides the closest architectural basis for LAD-COD because it derives a target embedding from the image-instruction pair to condition mask prediction.

Transferring this language-to-mask design to COD requires target guidance to work together with dense visual features that retain weak boundaries and subtle local variations. In representative LMM-to-SAM systems, however, the target embedding mainly serves as a sparse prompt for mask decoding, leaving top-down target guidance only loosely coupled with the bottom-up visual representation. LAD-COD addresses this disconnect through Language-Aligned Dual Visual Fusion (LADVF), which extends the target embedding beyond sparse prompting to retrieve patch-level language-aligned features and regulate their residual integration with the hierarchical representation.

%% file: sections/method.tex
\section{Methodology}
\label{sec:method}

\subsection{LAD-COD Framework}

We propose \textbf{LAD-COD}, a language-guided framework that aligns an instruction-conditioned target embedding with hierarchical dense visual evidence for camouflaged object detection. LAD-COD draws that embedding from the hidden state of a special \texttt{[SEG]} token and uses it to condition a SAM-based mask decoder, following language-to-mask design \cite{lai2024lisa}. Given an image $I$ and an instruction $T$, the LMM generates a textual response $\hat{y}$ alongside the token, and LAD-COD predicts a binary mask $\hat{M}$ for the camouflaged target that the image-instruction pair specifies.

Transferring the language-to-mask design to COD is non-trivial, as semantic conditioning must be combined with dense visual evidence that preserves weak boundaries, local texture variations, and low-contrast structures. LAD-COD addresses this requirement with a trainable hierarchical visual branch that learns camouflage-sensitive features directly from COD data, without fully adapting a large foundation image encoder. Hierarchical features preserve fine structural evidence, yet they lack explicit alignment with the target embedding that conditions mask prediction. Language-Aligned Dual Visual Fusion (LADVF) establishes this missing alignment by extending the target embedding beyond sparse prompting to query frozen patch-level language-aligned features and regulate their residual integration with the hierarchical representation, as illustrated in Figure~\ref{fig:pipeline}.

\subsubsection{Sparse Target Prompting}
From the image-instruction pair, the LMM produces the hidden state $h_t$ of the \texttt{[SEG]} token as the instruction-conditioned target representation. A lightweight projection maps $h_t$ into SAM's prompt embedding space as
\begin{equation}
    z = \phi_{\mathrm{tp}}(h_t),
\end{equation}
where $z$ is the projected target embedding. The SAM prompt encoder then maps $z$ to
\begin{equation}
    P_{\text{sparse}}, P_{\text{base}} =
    \operatorname{PromptEncoder}(z),
\end{equation}
where $P_{\text{sparse}}$ conditions the mask decoder with the target representation and $P_{\text{base}}$ is SAM's default dense prompt. Beyond sparse prompting, $z$ is subsequently reused by LADVF to retrieve language-aligned visual features for integration with the hierarchical representation.

\subsubsection{Hierarchical Visual Branch}
Reliable performance in COD depends on dense visual features that preserve weak boundaries and fine local structure while incorporating broad contextual information~\cite{Fan_2020_CVPR, Sun_2022}. We therefore adopt Swin Transformer~\cite{liu2021swin} as a SAM-compatible trainable hierarchical visual branch. The branch produces four stage-wise features $F_1, F_2, F_3, F_4$, whose spatial resolution decreases as contextual information is progressively aggregated. Each $F_i$ is bilinearly resized to the resolution of $F_1$ and mapped to 64 channels through a stage-specific $1\times1$ convolution; the aligned features are then concatenated as
\begin{equation}
    F_s = \operatorname{Concat}(
    \psi_1(F_1), \psi_2(F_2), \psi_3(F_3), \psi_4(F_4)),
\end{equation}
yielding $F_s \in \mathbb{R}^{B \times C_s \times H_s \times W_s}$ with $C_s=256$. Although $F_s$ retains the dense evidence required by COD, it is constructed independently of $z$ and therefore lacks explicit target alignment, which LADVF addresses.

\subsubsection{Mask Decoding}
After LADVF produces the fused feature $F_{\text{fused}}$ (described in the following subsection), it is projected to the image embedding expected by the SAM mask decoder,
\begin{equation}
    E_{\text{img}} = \phi_{\mathrm{img}}(F_{\text{fused}}),
\end{equation}
where $\phi_{\mathrm{img}}(\cdot)$ resizes its input to $64\times64$ and maps it to 256 channels. A learnable dense-prompt projection further derives
\begin{equation}
    P_{\text{dense}} = \phi_{\mathrm{dp}}(E_{\text{img}}),
\end{equation}
replacing the default $P_{\text{base}}$. The final mask is then predicted as
\begin{equation}
    \hat{M} =
    \operatorname{Decoder}(E_{\text{img}}, P_{\text{sparse}}, P_{\text{dense}}).
\end{equation}

\begin{table*}[!t]
\centering
\small
\setlength{\tabcolsep}{5pt}
\renewcommand{\arraystretch}{1.28}
\begin{tabular}{lcccc|cccc|cccc}
\hline
\multirow{2}{*}{Method}
& \multicolumn{4}{c}{CAMO} & \multicolumn{4}{c}{COD10K} & \multicolumn{4}{c}{NC4K} \\
\cline{2-13}
& $S_m\uparrow$ & $E_\phi\uparrow$ & $F_\beta^w\uparrow$ & MAE$\downarrow$
& $S_m\uparrow$ & $E_\phi\uparrow$ & $F_\beta^w\uparrow$ & MAE$\downarrow$
& $S_m\uparrow$ & $E_\phi\uparrow$ & $F_\beta^w\uparrow$ & MAE$\downarrow$ \\
\hline
SINet~\cite{Fan_2020_CVPR} & 0.745 & 0.804 & 0.644 & 0.092 & 0.776 & 0.864 & 0.631 & 0.043 & 0.808 & 0.871 & 0.723 & 0.058 \\
UGTR~\cite{yang2021ugtr} & 0.785 & 0.823 & 0.686 & 0.086 & 0.818 & 0.853 & 0.667 & 0.035 & 0.839 & 0.874 & 0.747 & 0.052 \\
C2FNet~\cite{chen2022c2fnet} & 0.796 & 0.854 & 0.719 & 0.080 & 0.813 & 0.869 & 0.686 & 0.036 & 0.837 & 0.874 & 0.762 & 0.052 \\
PreyNet~\cite{zhang2022preynet} & 0.790 & 0.842 & 0.708 & 0.077 & 0.813 & 0.881 & 0.697 & 0.034 & 0.834 & 0.887 & 0.763 & 0.050 \\
SegMaR~\cite{jia2022segmar} & 0.815 & 0.874 & 0.753 & 0.071 & 0.833 & 0.899 & 0.724 & 0.034 & 0.841 & 0.896 & 0.781 & 0.046 \\
BGNet~\cite{Sun_2022} & 0.816 & 0.871 & 0.751 & 0.069 & 0.831 & 0.901 & 0.722 & 0.033 & 0.851 & 0.907 & 0.788 & 0.044 \\
ZoomNet~\cite{pang2022zoom} & 0.820 & 0.877 & 0.752 & 0.066 & 0.838 & 0.888 & 0.729 & 0.029 & 0.853 & 0.896 & 0.784 & 0.043 \\
SAM-Adapter~\cite{chen2023samadapter} & 0.847 & 0.873 & 0.765 & 0.070 & 0.883 & 0.918 & 0.801 & 0.025 & 0.884 & 0.904 & 0.813 & 0.042 \\
HitNet~\cite{hu2023hitnet} & 0.849 & 0.906 & 0.809 & 0.055 & 0.871 & 0.935 & 0.806 & 0.023 & 0.875 & 0.926 & 0.834 & 0.037 \\
FEDER~\cite{he2023feder} & 0.802 & 0.867 & 0.738 & 0.071 & 0.822 & 0.900 & 0.716 & 0.032 & 0.846 & 0.905 & 0.789 & 0.045 \\
Camouflageator~\cite{he2023camouflageator} & 0.829 & 0.891 & 0.805 & 0.066 & 0.843 & 0.920 & 0.763 & 0.028 & 0.869 & 0.922 & 0.835 & 0.041 \\
CamoFormer~\cite{yin2024camoformer} & \underline{0.872} & 0.929 & 0.831 & 0.046 & 0.869 & 0.930 & 0.786 & 0.023 & 0.892 & 0.938 & 0.847 & 0.030 \\
ESCNet~\cite{Ye_2025_ICCV} & 0.871 & \underline{0.934} & \underline{0.843} & \underline{0.044} & 0.873 & \underline{0.939} & 0.804 & \underline{0.021} & 0.892 & \underline{0.941} & \underline{0.859} & \underline{0.028} \\
MM-SAM~\cite{ren2025multi} & 0.863 & 0.901 & 0.782 & 0.059 & \underline{0.896} & 0.907 & \underline{0.808} & 0.023 & \underline{0.898} & 0.911 & 0.820 & 0.038 \\
\hline
LAD-COD & \textbf{0.899} & \textbf{0.941} & \textbf{0.875} & \textbf{0.034}
& \textbf{0.898} & \textbf{0.955} & \textbf{0.854} & \textbf{0.017}
& \textbf{0.907} & \textbf{0.947} & \textbf{0.879} & \textbf{0.026} \\
\hline
\end{tabular}
\caption{Quantitative comparison on CAMO, COD10K, and NC4K datasets. Methods are ordered by publication year. The best results are highlighted in bold, and the second-best results are underlined.}
\label{tab:sota}
\end{table*}

\subsection{Language-Aligned Dual Visual Fusion}
\label{sec:ladvf}

LADVF establishes this missing alignment by extending $z$ beyond sparse prompting to query patch-level language-aligned features and regulate their residual integration with $F_s$ through a learned gate.

The frozen CLIP vision tower first extracts a patch-level feature map
\begin{equation}
    F_c = \operatorname{CLIP}(I),
\end{equation}
where $F_c \in \mathbb{R}^{B \times C_c \times H_c \times W_c}$. Within LADVF, $F_c$ serves as the patch-level language-aligned representation queried by the target embedding, while the dense representation for mask decoding is formed by subsequently fusing the retrieved information with $F_s$.

To make $F_c$ compatible with both retrieval and spatial fusion, we project its channels and align its resolution with $F_s$ using
\begin{equation}
    \tilde{F}_c = \operatorname{Resize}(\phi_{\mathrm{c}}(F_c), H_s, W_s),
\end{equation}
where $\phi_{\mathrm{c}}(\cdot)$ maps $C_c$ channels to the retrieval width $C_m$. Cross-attention operates in this retrieval dimension, while $\phi_{\mathrm{sem}}(\cdot)$ maps the retrieved semantic feature to the dense feature dimension $C_s$ before fusion. Following channel projection and resizing, $\tilde{F}_c \in \mathbb{R}^{B \times C_m \times H_s \times W_s}$ shares the spatial dimensions of $F_s$. Flattening its spatial axes yields
\begin{equation}
    X_c = \operatorname{Flatten}(\tilde{F}_c),
\end{equation}
where $X_c \in \mathbb{R}^{B \times H_sW_s \times C_m}$ forms the keys and values for cross-attention. The target embedding $z$ is projected into the same retrieval space as a single query token,
\begin{equation}
    Q = \operatorname{Unsqueeze}(\phi_{\mathrm{q}}(z), 1),
\end{equation}
where $\phi_{\mathrm{q}}(\cdot)$ maps $z$ to $C_m$ channels and $Q \in \mathbb{R}^{B \times 1 \times C_m}$.

LADVF then applies multi-head cross-attention with $Q$ as the query and $X_c$ as the keys and values to obtain a target-related semantic representation and its corresponding patch-level attention weights,
\begin{equation}
    \begin{aligned}
        s_0, \bar{A} &=
        \operatorname{MHA}(Q, X_c, X_c), \\
        s &= \operatorname{Squeeze}(\operatorname{LN}(s_0), 1), \\
        A &= \operatorname{Reshape}(\bar{A}, H_s, W_s).
    \end{aligned}
\end{equation}
Here, $\operatorname{MHA}(\cdot)$ and $\operatorname{LN}(\cdot)$ denote multi-head cross-attention and layer normalization, respectively. The tensor $\bar{A} \in \mathbb{R}^{B \times 1 \times H_sW_s}$ contains the patch-level attention weights, which are reshaped into $A \in \mathbb{R}^{B \times 1 \times H_s \times W_s}$ to restore their spatial correspondence. The normalized output $s \in \mathbb{R}^{B \times C_m}$ summarizes the retrieved language-aligned information.

To recover a dense semantic representation, $A$ weights the aligned CLIP feature spatially, while $s$ is broadcast across all spatial positions,
\begin{equation}
    F_{\text{sem}} =
    \rho\left(\tilde{F}_c \odot A + s\right),
\end{equation}
where $A$ is broadcast across channels, $\odot$ denotes element-wise multiplication, and $\rho(\cdot)$ is a convolutional refinement block. The resulting $F_{\text{sem}} \in \mathbb{R}^{B \times C_m \times H_s \times W_s}$ combines the retrieved semantic summary $s$ with the patch-weighted CLIP features in a spatially aligned representation.

To integrate $F_{\text{sem}}$ with $F_s$ while preserving the weak structural evidence carried by the hierarchical branch, LADVF uses gated residual fusion,
\begin{equation}
    G = \sigma(\eta([F_s, F_{\text{sem}}])),
\end{equation}
\begin{equation}
    F_{\text{fused}} = F_s + G \odot \phi_{\mathrm{sem}}(F_{\text{sem}}),
\end{equation}
Here, $[\cdot,\cdot]$ denotes channel-wise concatenation, allowing the convolutional network $\eta(\cdot)$ to predict a spatially varying gate from both $F_s$ and $F_{\text{sem}}$. The sigmoid function $\sigma(\cdot)$ bounds $G$ between 0 and 1, while $\phi_{\mathrm{sem}}(\cdot)$ maps $F_{\text{sem}}$ from $C_m$ to $C_s$ to match the channel width of $F_s$ before residual integration. The identity path leaves $F_s$ unchanged, while $G$ controls the spatial contribution of the retrieved semantics without overwriting fine local structure in $F_s$.

\subsection{Training Objective}

Training jointly supervises the textual response and predicted mask through language generation and segmentation objectives. The language modeling applies standard autoregressive cross-entropy to the response tokens as
\begin{equation}
    \mathcal{L}_{\text{text}}
    = -\sum_t \log p(y_t \mid y_{<t}, I, T).
\end{equation}
Here, $y_t$ is the ground-truth response token at step $t$, while $y_{<t}$ contains the preceding response tokens. The probability $p(\cdot)$ is conditioned on the input image $I$, instruction $T$, and preceding tokens.

Mask supervision combines binary cross-entropy and Dice losses to penalize pixelwise errors and region mismatch,
\begin{equation}
    \mathcal{L}_{\text{mask}} =
    \lambda_{\text{bce}}\mathcal{L}_{\text{bce}}(M, \hat{M})
    +
    \lambda_{\text{dice}}\mathcal{L}_{\text{dice}}(M, \hat{M}).
\end{equation}
Here, $M$ and $\hat{M}$ denote the ground-truth and predicted masks, respectively. The coefficients $\lambda_{\text{bce}}$ and $\lambda_{\text{dice}}$ weight the two mask losses.

The complete training objective combines the language-generation and mask-supervision terms as
\begin{equation}
    \mathcal{L} =
    \lambda_{\text{text}}\mathcal{L}_{\text{text}}
    +
    \mathcal{L}_{\text{mask}}.
\end{equation}
The coefficient $\lambda_{\text{text}}$ balances language generation against mask supervision during joint optimization. Following LISA~\cite{lai2024lisa}, we use $\lambda_{\text{text}}=1.0$, $\lambda_{\text{bce}}=2.0$, and $\lambda_{\text{dice}}=0.5$ in all experiments.

\begin{figure*}[!t]
    \centering
    \includegraphics[width=\textwidth]{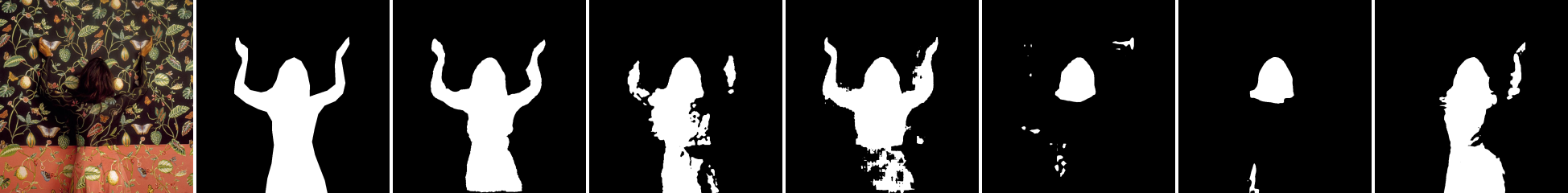}\\[0.04em]
    \includegraphics[width=\textwidth]{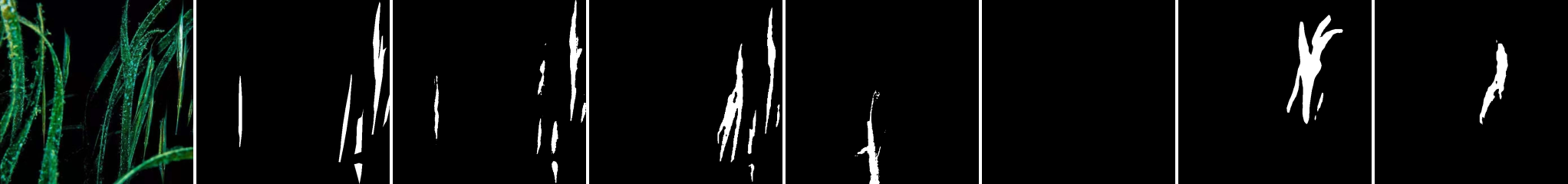}\\[0.04em]
    \includegraphics[width=\textwidth]{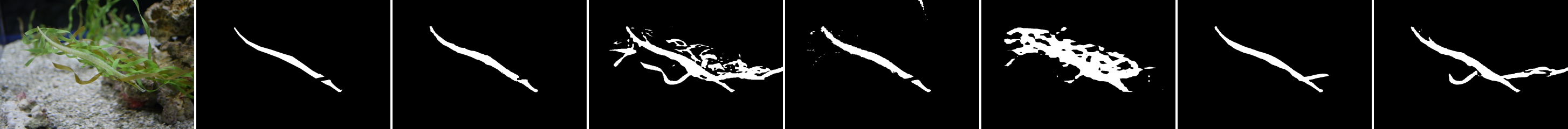}\\[0.04em]
    \includegraphics[width=\textwidth]{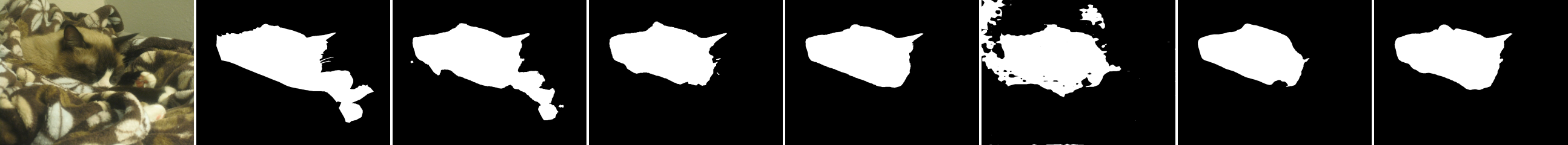}\\[0.04em]
    \includegraphics[width=\textwidth]{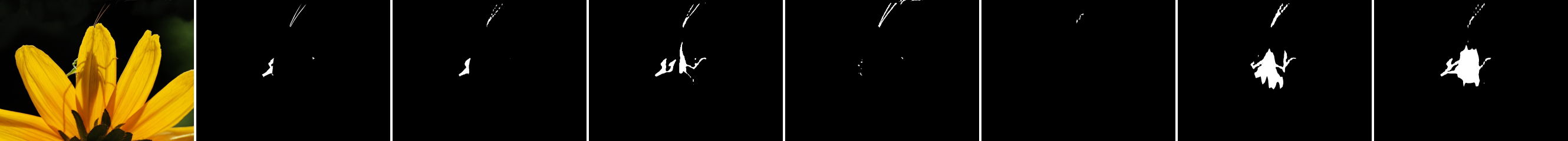}\\[0.04em]
    \includegraphics[width=\textwidth]{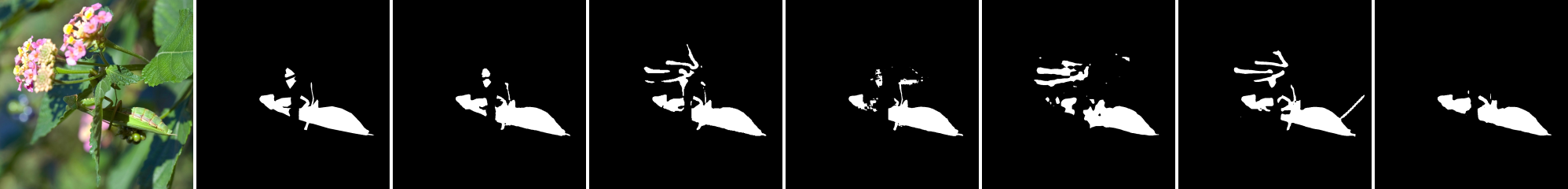}\\[0.1em]
    {\small
    \begin{minipage}[t]{0.125\textwidth}\centering Image\end{minipage}%
    \begin{minipage}[t]{0.125\textwidth}\centering GT\end{minipage}%
    \begin{minipage}[t]{0.125\textwidth}\centering Ours\end{minipage}%
    \begin{minipage}[t]{0.125\textwidth}\centering ESCNet\end{minipage}%
    \begin{minipage}[t]{0.125\textwidth}\centering MM-SAM\end{minipage}%
    \begin{minipage}[t]{0.125\textwidth}\centering UGTR\end{minipage}%
    \begin{minipage}[t]{0.125\textwidth}\centering HitNet\end{minipage}%
    \begin{minipage}[t]{0.125\textwidth}\centering CamoFormer\end{minipage}%
    }
    \caption{Qualitative comparison of different methods on challenging COD samples.}
    \label{fig:qualitative_comparison}
\end{figure*}

%% file: sections/experiment.tex
\section{Experiments}

\subsection{Experimental Setup}

\subsubsection{Datasets.}
We conduct experiments on three standard COD benchmarks, namely CAMO~\cite{le2019anabranch}, COD10K~\cite{Fan_2020_CVPR}, and NC4K~\cite{lv2021ranknet}. CAMO contains 1,000 training and 250 test images, COD10K contains 3,040 training and 2,026 test images, and NC4K provides 4,121 test images. Following the established COD protocol~\cite{Fan_2020_CVPR, ren2025multi, Ye_2025_ICCV}, we train on the combined CAMO and COD10K training sets and evaluate on all three test sets.

\subsubsection{Evaluation Metrics.}
We report structure-measure $S_m$~\cite{fan2017structure}, mean E-measure $E_\phi$~\cite{fan2018enhanced}, weighted F-measure $F_\beta^w$~\cite{margolin2014evaluate}, and mean absolute error (MAE). The $S_m$ score measures structural agreement with the ground truth, whereas $E_\phi$ combines local pixel agreement with global foreground statistics.  $F_\beta^w$ weights precision-recall errors by spatial context, while MAE records the mean pixel-wise difference between the prediction and ground truth.

\subsubsection{Implementation Details.}
LAD-COD uses LISA-7B as the LMM backbone, the SAM mask decoder, and Swin-L~\cite{liu2021swin} as its trainable hierarchical visual branch. Swin-L is initialized from the window-12, $384\times384$ checkpoint, and input images are resized to $1024\times1024$ and paired with a fixed instruction. Training uses four NVIDIA RTX 3090 GPUs with DeepSpeed, a per-GPU batch size of 1 and a gradient accumulation over 10 steps, giving an effective batch size of 40. We use AdamW with an initial learning rate of $1\times10^{-4}$, while the Swin branch uses  $0.1\times$ this rate. The model is trained for 10 epochs under the joint language-generation and mask-supervision objective defined in the Method section. Following LISA~\cite{lai2024lisa}, we apply LoRA mainly to the language model's query and value projections while keeping most LMM parameters frozen.

\subsection{Comparison with State-of-the-Art Methods}

\subsubsection{Quantitative Comparison.}
Table~\ref{tab:sota} compares LAD-COD with representative conventional, Transformer-based, and foundation-model approaches on the three common COD test sets. LAD-COD achieves the best performance on all evaluation metrics, reflecting improved foreground completeness and reduced pixel-level errors. Its consistent superiority across all three datasets demonstrates strong generalization across varying scene complexities, object appearances, and levels of background distractions.

\subsubsection{Qualitative Comparison.}
Figure~\ref{fig:qualitative_comparison} compares LAD-COD with representative methods under clutter, weak boundaries, small targets, and similar foreground-background textures. Competing predictions omit thin structures, fragment targets, or respond to similar backgrounds, whereas LAD-COD recovers more complete masks with fewer background responses around narrow appendages and low-contrast contours, better preserving the fine structures that are most easily lost in camouflage scenes.

\begin{table*}[!t]
\centering
\small
\setlength{\tabcolsep}{5pt}
\renewcommand{\arraystretch}{1.20}
\begin{tabular}{l|cccc|cccc|cccc}
\hline
\multirow{2}{*}{Variant}
& \multicolumn{4}{c|}{CAMO} & \multicolumn{4}{c|}{COD10K} & \multicolumn{4}{c}{NC4K} \\
\cline{2-13}
& $S_m\uparrow$ & $E_\phi\uparrow$ & $F_\beta^w\uparrow$ & MAE$\downarrow$
& $S_m\uparrow$ & $E_\phi\uparrow$ & $F_\beta^w\uparrow$ & MAE$\downarrow$
& $S_m\uparrow$ & $E_\phi\uparrow$ & $F_\beta^w\uparrow$ & MAE$\downarrow$ \\
\hline
Swin-SAM & 0.816 & 0.871 & 0.681 & 0.080 & 0.778 & 0.828 & 0.577 & 0.048 & 0.814 & 0.872 & 0.670 & 0.065 \\
+ target embedding & 0.889 & 0.937 & 0.857 & 0.039 & 0.879 & 0.946 & 0.815 & 0.020 & 0.898 & 0.945 & 0.865 & 0.028 \\
+ four-stage features & 0.895 & 0.937 & 0.864 & 0.036 & 0.890 & 0.948 & 0.835 & \textbf{0.018} & 0.904 & 0.944 & 0.872 & \textbf{0.026} \\
+ LADVF (LAD-COD) & \textbf{0.897} & \textbf{0.942} & \textbf{0.869} & \textbf{0.035} & \textbf{0.891} & \textbf{0.952} & \textbf{0.840} & \textbf{0.018} & \textbf{0.905} & \textbf{0.948} & \textbf{0.876} & \textbf{0.026} \\
\hline
\end{tabular}
\caption{Cumulative ablation of the target embedding, four-stage features, and LADVF at $768\times768$ resolution.}
\label{tab:ablation}
\end{table*}

\begin{figure*}[!t]
    \centering
    \includegraphics[width=\textwidth]{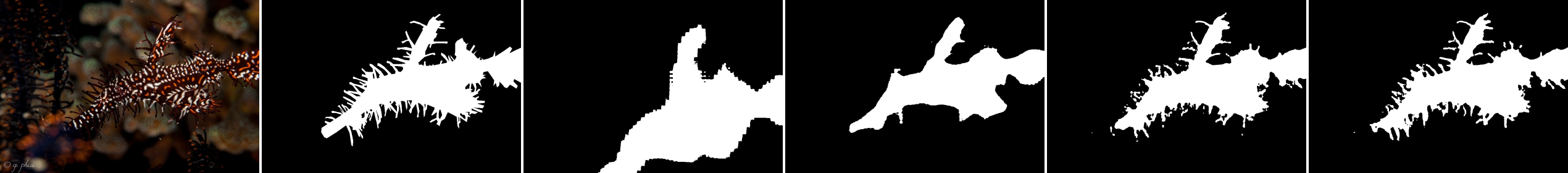}\\[0.02em]
    \includegraphics[width=\textwidth]{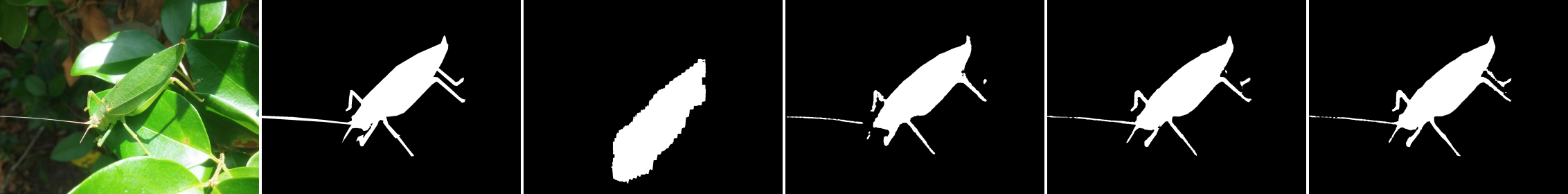}\\[0.02em]
    \includegraphics[width=\textwidth]{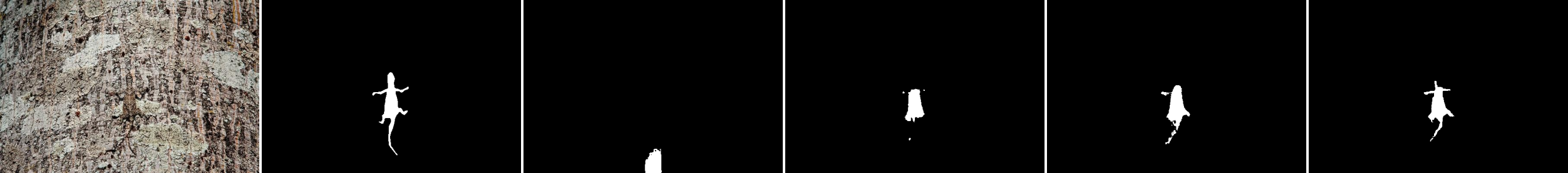}\\[0.02em]
    \includegraphics[width=\textwidth]{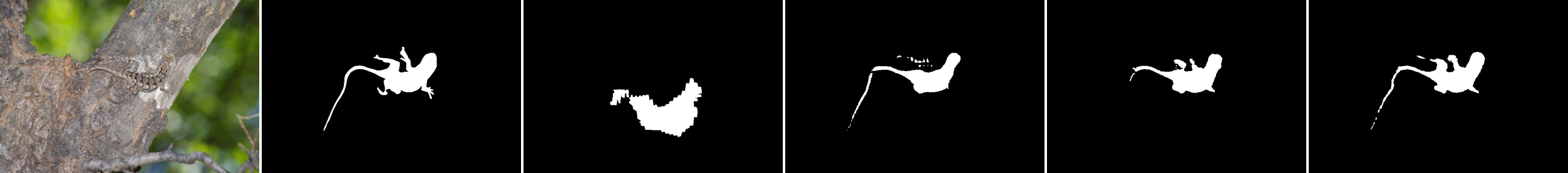}\\[0.02em]
    \includegraphics[width=\textwidth]{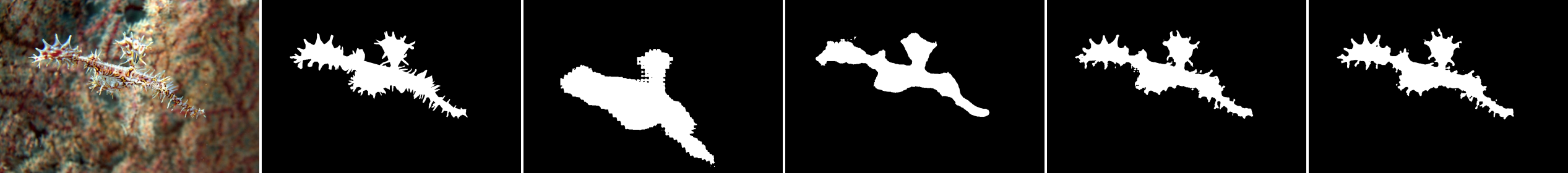}\\[0.02em]
    \includegraphics[width=\textwidth]{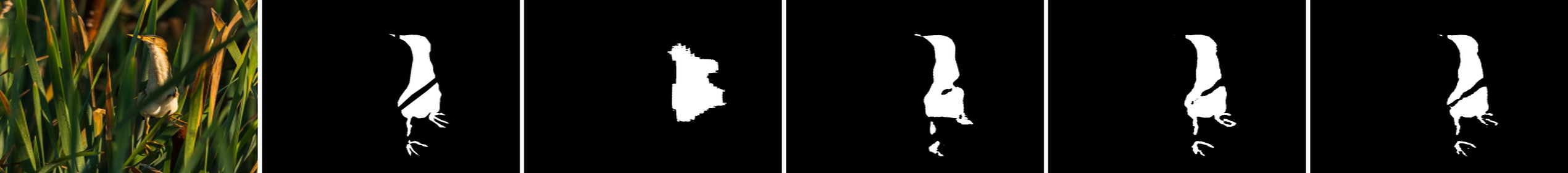}\\[0.05em]
    {\small
    \begin{minipage}[t]{0.166\textwidth}\centering Image\end{minipage}%
    \begin{minipage}[t]{0.166\textwidth}\centering GT\end{minipage}%
    \begin{minipage}[t]{0.166\textwidth}\centering Swin-SAM\end{minipage}%
    \begin{minipage}[t]{0.166\textwidth}\centering $+$ target emb.\end{minipage}%
    \begin{minipage}[t]{0.166\textwidth}\centering $+$ four-stage\end{minipage}%
    \begin{minipage}[t]{0.166\textwidth}\centering $+$ LADVF\end{minipage}%
    }
    \caption{Qualitative ablation results on challenging camouflaged scenes.}
    \label{fig:ablation_visualization}
\end{figure*}

\subsection{Ablation Study}

We assess the contributions of the target embedding, four-stage features, and LADVF through cumulative ablations. All variants share the same visual backbone, mask decoder, $768\times768$ input resolution, optimization settings, and training duration, with results reported in Table~\ref{tab:ablation}. 

\subsubsection{Swin-SAM Baseline.}
To establish the reference configuration, Swin-SAM uses only Swin's final-stage feature in the dense image pathway and omits the instruction-conditioned target embedding and LADVF. Its lower performance across all three datasets indicates the limitations of the final-stage representation when target conditioning, multi-stage structure, and alignment-based fusion are absent.

\subsubsection{Target Embedding.}
To evaluate instruction-conditioned localization, we add the target embedding as a sparse prompt while retaining the final-stage dense feature. This yields the largest improvement across all three datasets. Because the dense pathway remains unchanged, the result isolates top-down target guidance as the main localization gain and helps distinguish the target from similar backgrounds. The improvement therefore reflects better target selection rather than a change in visual encoding.

\subsubsection{Four-Stage Features.}
With target conditioning fixed, we replace the final-stage representation with four-stage features to evaluate the contribution of multi-level structure. The resulting gains indicate that combining early-stage features with later-stage representations complements target guidance, recovering weak boundaries and thin structures under low foreground-background contrast.

\subsubsection{LADVF.}
With target conditioning and hierarchical features in place, the final variant adds LADVF to evaluate their explicit alignment. Its smaller but consistent gains show that feature-level alignment complements the preceding localization and structural improvements.

Figure~\ref{fig:ablation_visualization} shows the target embedding expanding coarse masks, four-stage features restoring weak structures, and LADVF refining residual gaps, consistent with Table~\ref{tab:ablation}.

%% file: sections/conclusion.tex
\section{Conclusion}

This work presents LAD-COD, a Language-Aligned Dense perception framework for COD, which integrates a trainable hierarchical visual branch with Language-Aligned Dual Visual Fusion (LADVF). The instruction-conditioned target embedding provides top-down target guidance for mask decoding, while the hierarchical branch provides bottom-up visual features that retain low-contrast boundaries, fine textures, and multi-scale structure. LADVF connects these two by querying language-aligned visual semantics with the target embedding, then injecting the retrieved semantics through gated residual fusion. This design improves localization while retaining the visual details needed for camouflage segmentation. Experiments on CAMO, COD10K, and NC4K demonstrate its effectiveness, and controlled ablations clarify the roles of target prompting, multi-stage features, and LADVF. More broadly, our results suggest that language is useful for COD not as a substitute for visual detail, but as a means of organizing weak visual cues around the target. Future work will explore efficient fusion under diverse referring expressions and multi-object scenes.


%% file: sections/supplement.tex
\clearpage
\appendix
\setcounter{secnumdepth}{1}
\section*{Appendix}
\addcontentsline{toc}{section}{Appendix}

\noindent\textbf{1.\ Additional Implementation Details.}
\label{sec:supp-impl}

Table~\ref{tab:supp-config} summarizes the main training configuration of LAD-COD.
The CLIP~\cite{radford2021learning} vision tower, the LLaVA~\cite{liu2024visual} multimodal projector, the SAM~\cite{kirillov2023segment} prompt encoder, and most LLM weights remain frozen; trainable modules include LoRA adapters, \texttt{embed\_tokens}, \texttt{lm\_head}, \texttt{text\_hidden\_fcs}, the SAM mask decoder, and the hierarchical Swin~\cite{liu2021swin} encoder with LADVF, following the LISA~\cite{lai2024lisa} training paradigm.

\vspace{1.25em}
\noindent
\begin{minipage}{\columnwidth}
\centering
\small
\setlength{\tabcolsep}{4pt}
\renewcommand{\arraystretch}{1.18}
\begin{tabular}{ll}
\toprule
Component & Value \\
\midrule
LMM backbone & LISA-7B \\
Hierarchical branch & Swin-L (window-12, $384$ pretrain) \\
Mask decoder & SAM \\
CLIP tower & ViT-L/14 (frozen) \\
Input resolution & $1024\times1024$ \\
\midrule
Optimizer & AdamW \\
Base learning rate & $1\times10^{-4}$ \\
Swin/LADVF learning rate & $0.1\times$ base \\
LoRA rank / $\alpha$ / dropout & $8$ / $16$ / $0.05$ \\
LoRA targets & \texttt{q\_proj}, \texttt{v\_proj} \\
\midrule
GPUs & $4\times$ RTX 3090 \\
DeepSpeed & Yes \\
Batch size (per GPU) & $1$ \\
Gradient accumulation & $10$ \\
Effective batch size & $40$ \\
Training epochs & $10$ \\
\midrule
Loss $\lambda_{\text{text}}$ & $1.0$ \\
Loss $\lambda_{\text{bce}}$ & $2.0$ \\
Loss $\lambda_{\text{dice}}$ & $0.5$ \\
\midrule
LADVF width $C_m$ & $128$ \\
LADVF attention heads & $8$ \\
Concatenated Swin width $C_s$ & $256$ ($64$ per stage) \\
Gate bias init & $-2$ \\
\bottomrule
\end{tabular}
\captionof{table}{Training configuration for LAD-COD.}
\label{tab:supp-config}
\end{minipage}

\vspace{1.45em}
\noindent\textbf{Instruction Templates.}
Training and evaluation randomly sample one of three equivalent instructions:
\begin{itemize}\setlength{\itemsep}{0.45em}\setlength{\parsep}{0pt}\setlength{\topsep}{0.65em}\setlength{\partopsep}{0pt}
    \item \texttt{Can you segment the camouflaged object in this image?}
    \item \texttt{Please segment the camouflaged object in this image.}
    \item \texttt{What is the camouflaged object in this image? Please respond with segmentation mask.}
\end{itemize}

\vspace{1.05em}
\noindent\textbf{Response Templates.}
Following LISA~\cite{lai2024lisa}, the supervised textual response is randomly sampled from five short templates that all contain the \texttt{[SEG]} token:
\begin{itemize}\setlength{\itemsep}{0.45em}\setlength{\parsep}{0pt}\setlength{\topsep}{0.65em}\setlength{\partopsep}{0pt}
    \item \texttt{It is [SEG].}
    \item \texttt{Sure, [SEG].}
    \item \texttt{Sure, it is [SEG].}
    \item \texttt{Sure, the segmentation result is [SEG].}
    \item \texttt{[SEG].}
\end{itemize}
The corresponding \texttt{[SEG]} hidden state is used as the target embedding for mask decoding.

\noindent\textbf{2.\ Additional Experiments.}
\label{sec:supp-extra-exp}

We further study the sensitivity of LAD-COD to the learning rate, input resolution, and Swin backbone capacity.
Unless otherwise stated, all variants follow the same LAD-COD training protocol as in the main paper, and are evaluated on CAMO~\cite{le2019anabranch}, COD10K~\cite{Fan_2020_CVPR}, and NC4K~\cite{lv2021ranknet} with $S_m$~\cite{fan2017structure}, $E_\phi$~\cite{fan2018enhanced}, $F_\beta^w$~\cite{margolin2014evaluate}, and MAE.

\vspace{1.05em}
\noindent\textbf{Learning-Rate Sensitivity.}
\label{sec:supp-lr}
We study learning-rate sensitivity under a compact setting: the full LAD-COD framework is retained, but the hierarchical branch is replaced with Swin-T.
All runs use $768\times768$ inputs and differ only in the base learning rate, chosen from $\{3\times10^{-4},\,1\times10^{-4},\,3\times10^{-5}\}$.
Table~\ref{tab:supp-lr} shows that $1\times10^{-4}$ performs best across datasets, whereas $3\times10^{-4}$ is clearly worse.
The smaller rate $3\times10^{-5}$ collapses in late training and yields much lower final metrics.

\vspace{1.05em}
\noindent
\begin{minipage}{\columnwidth}
\centering
\small
\setlength{\tabcolsep}{5pt}
\renewcommand{\arraystretch}{1.18}
\begin{tabular}{llcccc}
\hline
Learning rate & Dataset & $S_m\uparrow$ & $E_\phi\uparrow$ & $F_\beta^w\uparrow$ & MAE$\downarrow$ \\
\hline
\multirow{3}{*}{$3\times10^{-4}$}
& CAMO & 0.772 & 0.800 & 0.665 & 0.119 \\
& COD10K & 0.803 & 0.865 & 0.665 & 0.051 \\
& NC4K & 0.793 & 0.831 & 0.672 & 0.093 \\
\hline
\multirow{3}{*}{$1\times10^{-4}$}
& CAMO & \textbf{0.836} & \textbf{0.869} & \textbf{0.765} & \textbf{0.075} \\
& COD10K & \textbf{0.853} & \textbf{0.918} & \textbf{0.764} & \textbf{0.029} \\
& NC4K & \textbf{0.864} & \textbf{0.908} & \textbf{0.801} & \textbf{0.047} \\
\hline
\multirow{3}{*}{$3\times10^{-5}$}
& CAMO & 0.496 & 0.550 & 0.385 & 0.285 \\
& COD10K & 0.587 & 0.686 & 0.391 & 0.132 \\
& NC4K & 0.549 & 0.627 & 0.433 & 0.227 \\
\hline
\end{tabular}
\captionof{table}{Learning-rate sensitivity of LAD-COD with Swin-T at $768\times768$ resolution.}
\label{tab:supp-lr}
\end{minipage}

\vspace{1.0em}
\noindent\textbf{Input-Resolution Sensitivity.}
\label{sec:supp-res}
With Swin-L under the full LAD-COD protocol, we vary the input resolution over $\{384,\,512,\,768,\,1024\}$.
Table~\ref{tab:supp-res} shows steady gains from $384\times384$ to $768\times768$, with a smaller further gain at $1024\times1024$, which remains best on most metrics.

\noindent
\begin{minipage}{\columnwidth}
\centering
\small
\setlength{\tabcolsep}{5pt}
\renewcommand{\arraystretch}{1.18}
\begin{tabular}{llcccc}
\hline
Resolution & Dataset & $S_m\uparrow$ & $E_\phi\uparrow$ & $F_\beta^w\uparrow$ & MAE$\downarrow$ \\
\hline
\multirow{3}{*}{$384\times384$}
& CAMO & 0.869 & 0.928 & 0.827 & 0.045 \\
& COD10K & 0.842 & 0.925 & 0.753 & 0.026 \\
& NC4K & 0.873 & 0.932 & 0.828 & 0.034 \\
\hline
\multirow{3}{*}{$512\times512$}
& CAMO & 0.882 & 0.937 & 0.850 & 0.039 \\
& COD10K & 0.868 & 0.943 & 0.806 & 0.021 \\
& NC4K & 0.888 & 0.941 & 0.855 & 0.030 \\
\hline
\multirow{3}{*}{$768\times768$}
& CAMO & 0.897 & \textbf{0.942} & 0.869 & 0.035 \\
& COD10K & 0.891 & 0.952 & 0.840 & 0.018 \\
& NC4K & 0.905 & \textbf{0.948} & 0.876 & \textbf{0.026} \\
\hline
\multirow{3}{*}{$1024\times1024$}
& CAMO & \textbf{0.900} & 0.941 & \textbf{0.875} & \textbf{0.034} \\
& COD10K & \textbf{0.898} & \textbf{0.955} & \textbf{0.854} & \textbf{0.017} \\
& NC4K & \textbf{0.907} & 0.947 & \textbf{0.879} & \textbf{0.026} \\
\hline
\end{tabular}
\captionof{table}{Input-resolution sensitivity of LAD-COD (Swin-L).}
\label{tab:supp-res}
\end{minipage}

\noindent\textbf{Swin Backbone Variants.}
\label{sec:supp-swin}
We compare Swin-T/S/B/L~\cite{liu2021swin} under the same LAD-COD protocol at $768\times768$.
Table~\ref{tab:supp-swin} shows a clear capacity trend---Swin-T is weakest, Swin-S and Swin-B are close, and Swin-L is best on all metrics---supporting Swin-L as the default backbone.

\noindent
\begin{minipage}{\columnwidth}
\centering
\small
\setlength{\tabcolsep}{5pt}
\renewcommand{\arraystretch}{1.18}
\begin{tabular}{llcccc}
\hline
Backbone & Dataset & $S_m\uparrow$ & $E_\phi\uparrow$ & $F_\beta^w\uparrow$ & MAE$\downarrow$ \\
\hline
\multirow{3}{*}{Swin-T}
& CAMO & 0.836 & 0.869 & 0.765 & 0.075 \\
& COD10K & 0.853 & 0.918 & 0.764 & 0.029 \\
& NC4K & 0.864 & 0.908 & 0.801 & 0.047 \\
\hline
\multirow{3}{*}{Swin-S}
& CAMO & 0.858 & 0.902 & 0.807 & 0.055 \\
& COD10K & 0.863 & 0.929 & 0.786 & 0.025 \\
& NC4K & 0.879 & 0.928 & 0.832 & 0.036 \\
\hline
\multirow{3}{*}{Swin-B}
& CAMO & 0.859 & 0.907 & 0.808 & 0.056 \\
& COD10K & 0.868 & 0.936 & 0.799 & 0.023 \\
& NC4K & 0.880 & 0.928 & 0.835 & 0.036 \\
\hline
\multirow{3}{*}{Swin-L}
& CAMO & \textbf{0.897} & \textbf{0.942} & \textbf{0.869} & \textbf{0.035} \\
& COD10K & \textbf{0.891} & \textbf{0.952} & \textbf{0.840} & \textbf{0.018} \\
& NC4K & \textbf{0.905} & \textbf{0.948} & \textbf{0.876} & \textbf{0.026} \\
\hline
\end{tabular}
\captionof{table}{Comparison of Swin backbone variants in LAD-COD at $768\times768$ resolution.}
\label{tab:supp-swin}
\end{minipage}

\noindent\textbf{3.\ Target Embedding Visualization.}
\label{sec:supp-seg-focus}

Figure~\ref{fig:supp-seg-focus} visualizes the spatial focus induced by the target embedding on an NC4K~\cite{lv2021ranknet} sample and a COD10K~\cite{Fan_2020_CVPR} sample (Image / GT / focus map).
Part of the response lands on the camouflaged object, while other high responses appear on background regions with similar material and texture.
This suggests that the target embedding provides a coarse, target-related prior that helps pull attention toward the camouflaged region, but is not sufficient alone for precise segmentation---motivating the hierarchical visual branch, LADVF, and the mask decoder in LAD-COD.

\noindent\textbf{4.\ More Visualization Results.}
\label{sec:supp-qual}

This section provides additional qualitative examples beyond those in the main paper.
The samples are drawn from the CAMO~\cite{le2019anabranch}, COD10K~\cite{Fan_2020_CVPR}, and NC4K~\cite{lv2021ranknet} test sets, covering aquatic, terrestrial, and flying camouflaged scenes under clutter, weak boundaries, and look-alike textures.

\begin{center}
\includegraphics[width=\linewidth]{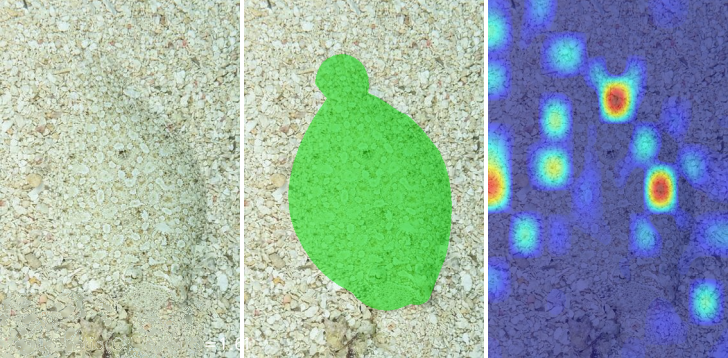}\\[0.35em]
\includegraphics[width=\linewidth]{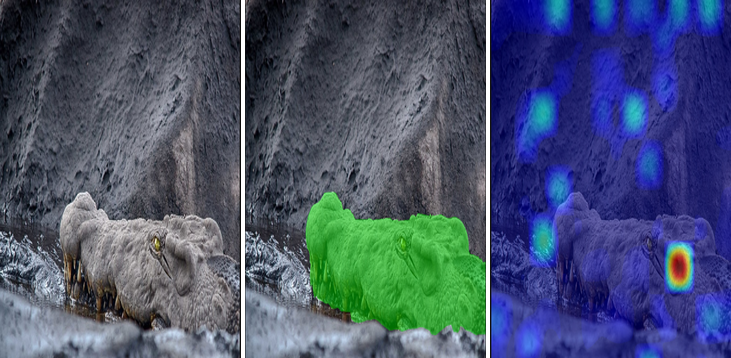}\\[0.2em]
{\footnotesize
\begin{minipage}[t]{0.32\linewidth}\centering Image\end{minipage}\hfill
\begin{minipage}[t]{0.32\linewidth}\centering GT\end{minipage}\hfill
\begin{minipage}[t]{0.32\linewidth}\centering Target embedding\end{minipage}%
}
\captionof{figure}{Target-embedding focus on NC4K and COD10K samples. Brighter regions indicate stronger target-embedding response.}
\label{fig:supp-seg-focus}
\end{center}

Figure~\ref{fig:supp-qual-cmp} shows more comparisons with ESCNet~\cite{Ye_2025_ICCV}, MM-SAM~\cite{ren2025multi}, UGTR~\cite{yang2021ugtr}, HitNet~\cite{hu2023hitnet}, and CamoFormer~\cite{yin2024camoformer} (Image / GT / Ours / baselines).
These extra samples are consistent with the main-paper comparisons: LAD-COD is usually closer to the ground truth, whereas competing methods more often miss the target, return fragmented masks, or respond to look-alike backgrounds.

Figure~\ref{fig:supp-qual-abl} presents further cumulative ablation examples under the same protocol as the main paper.
Swin-SAM uses only the final-stage Swin feature with a SAM decoder; each ``$+$'' column then cumulatively adds the target embedding, four-stage features, and LADVF.
The added cases again show progressive gains from localization to structural recovery and residual refinement.

Figure~\ref{fig:supp-qual} reports more Image / GT / LAD-COD masks.
Overall, LAD-COD can localize camouflaged targets and produce coherent object-level segmentations, often preserving the main body shape with a moderate amount of boundary detail.
Challenges remain when the object surface is partly occluded or the structure is highly fine-grained (e.g., thin branches and delicate appendages), where the predicted masks become coarser and miss fine parts.
In some cases, localization is also imperfect: similarly textured background regions are included, leading to over-segmentation beyond the true object extent.
For thin or occluded structures, stronger boundary-aware supervision may help recover missing fine parts; for look-alike background responses, harder negative mining on similarly textured regions may tighten target--background discrimination and reduce over-segmentation.

\clearpage
\onecolumn
\begin{center}
\includegraphics[width=\textwidth]{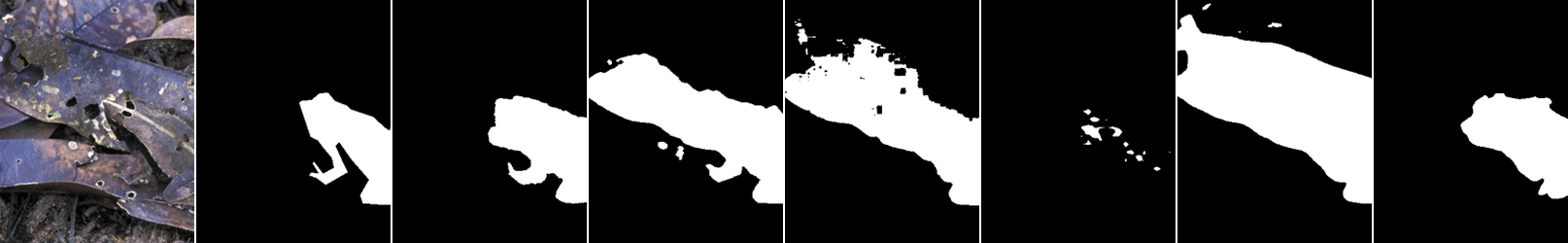}\\[0.03em]
\includegraphics[width=\textwidth]{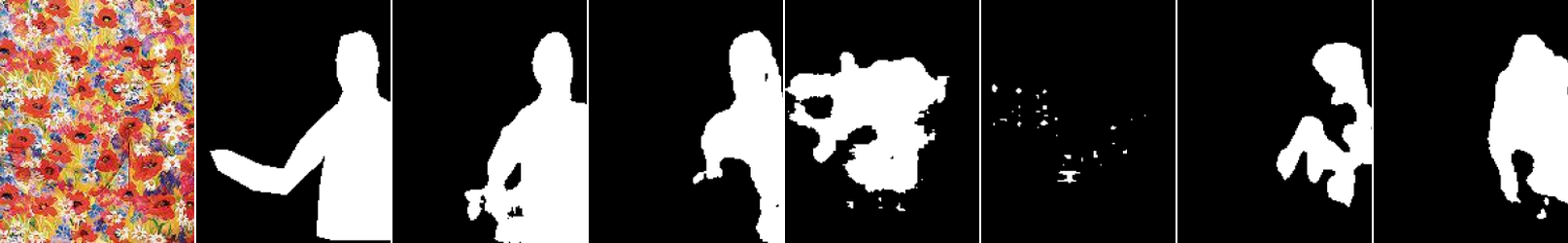}\\[0.03em]
\includegraphics[width=\textwidth]{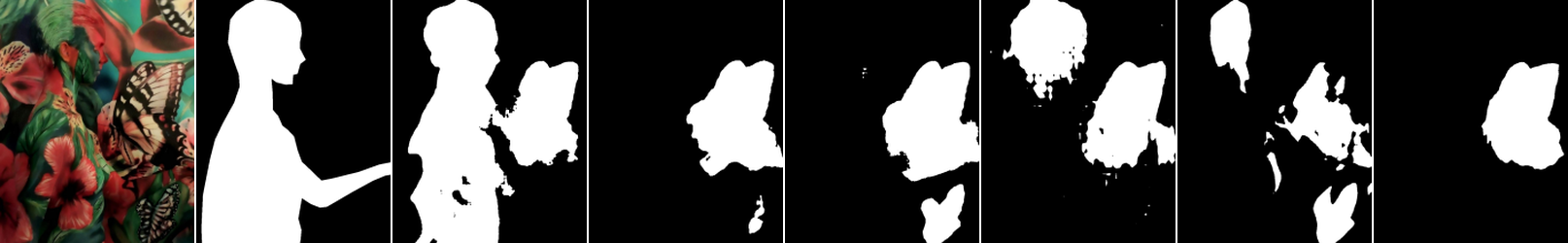}\\[0.03em]
\includegraphics[width=\textwidth]{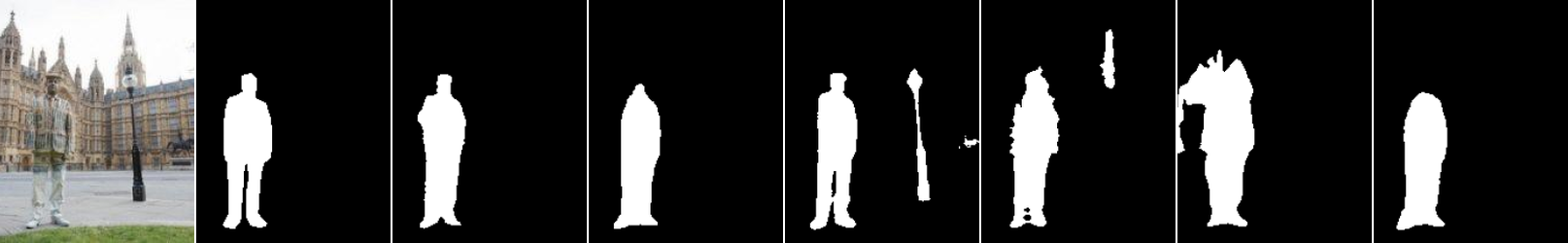}\\[0.03em]
\includegraphics[width=\textwidth]{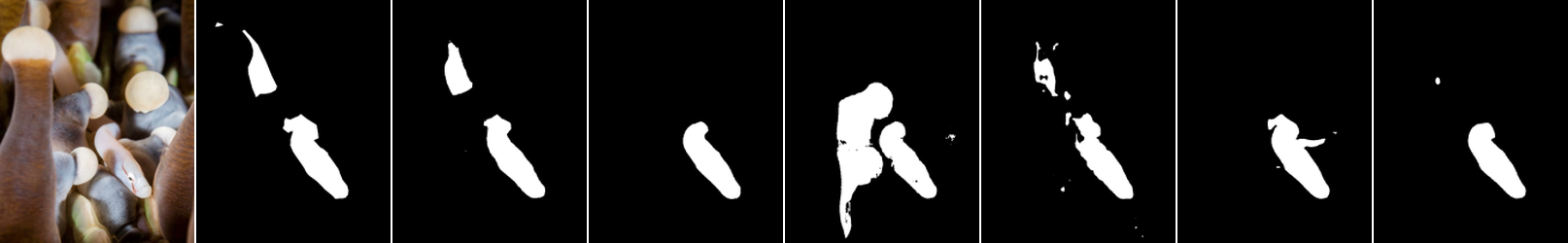}\\[0.03em]
\includegraphics[width=\textwidth]{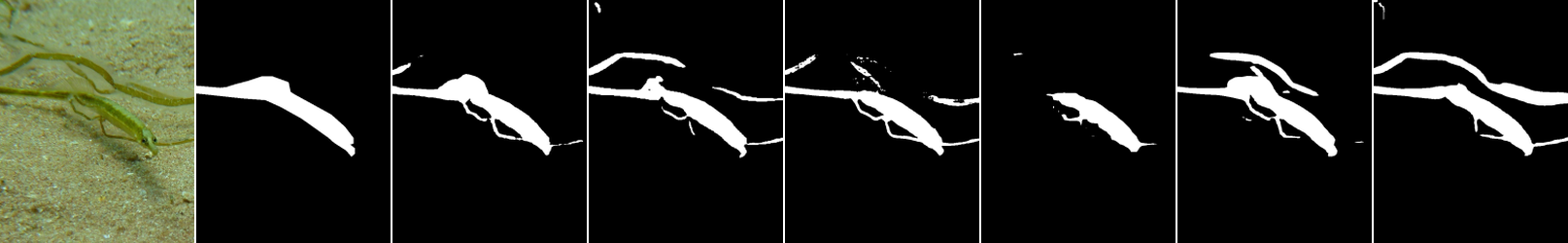}\\[0.03em]
\includegraphics[width=\textwidth]{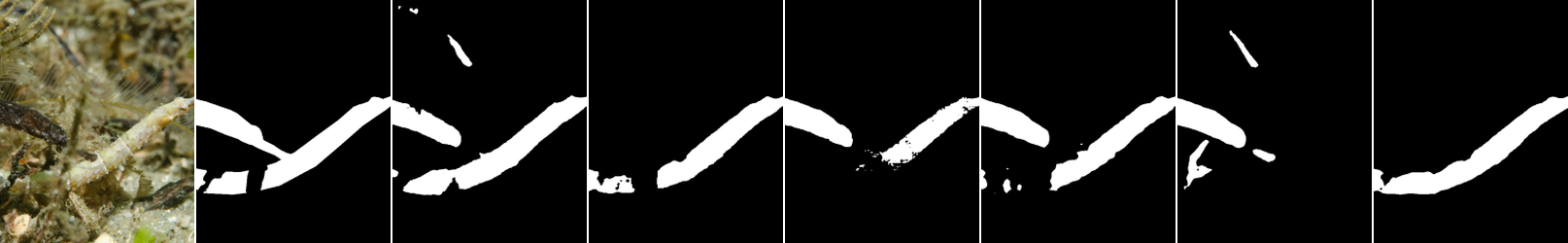}\\[0.08em]
{\small
\begin{minipage}[t]{0.125\textwidth}\centering Image\end{minipage}%
\begin{minipage}[t]{0.125\textwidth}\centering GT\end{minipage}%
\begin{minipage}[t]{0.125\textwidth}\centering Ours\end{minipage}%
\begin{minipage}[t]{0.125\textwidth}\centering ESCNet\end{minipage}%
\begin{minipage}[t]{0.125\textwidth}\centering MM-SAM\end{minipage}%
\begin{minipage}[t]{0.125\textwidth}\centering UGTR\end{minipage}%
\begin{minipage}[t]{0.125\textwidth}\centering HitNet\end{minipage}%
\begin{minipage}[t]{0.125\textwidth}\centering CamoFormer\end{minipage}%
}
\end{center}

\clearpage
\begin{center}
\includegraphics[width=\textwidth]{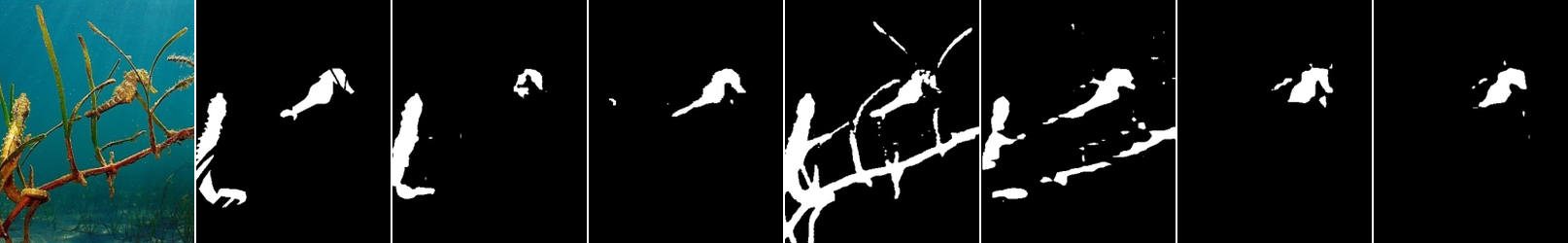}\\[0.03em]
\includegraphics[width=\textwidth]{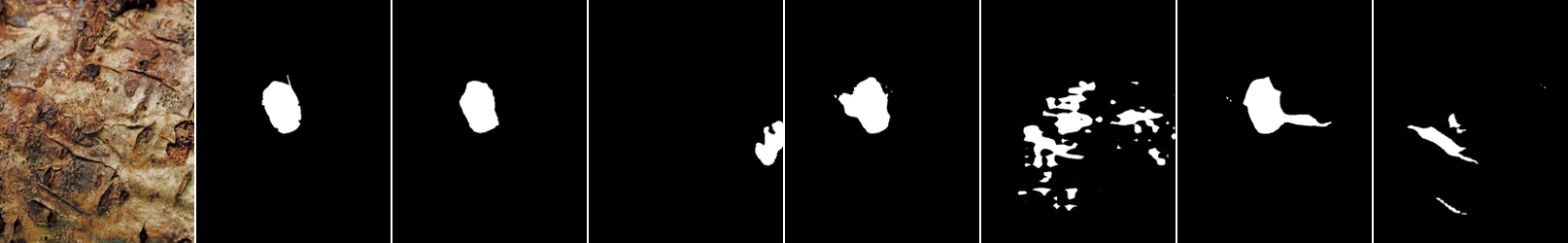}\\[0.03em]
\includegraphics[width=\textwidth]{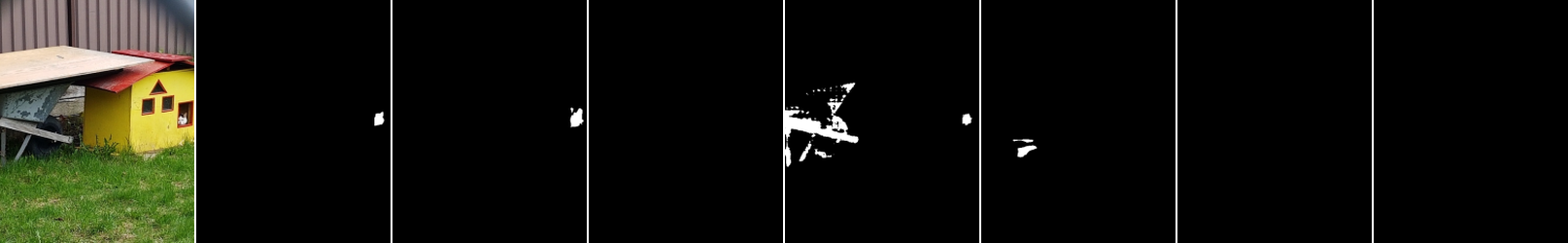}\\[0.03em]
\includegraphics[width=\textwidth]{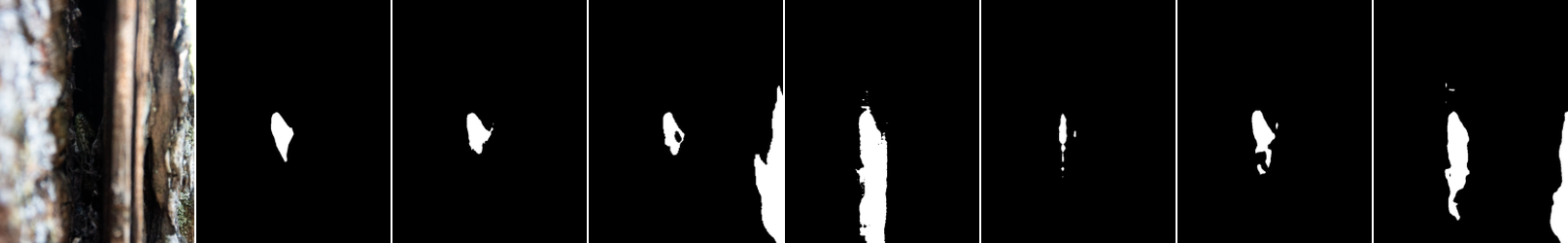}\\[0.03em]
\includegraphics[width=\textwidth]{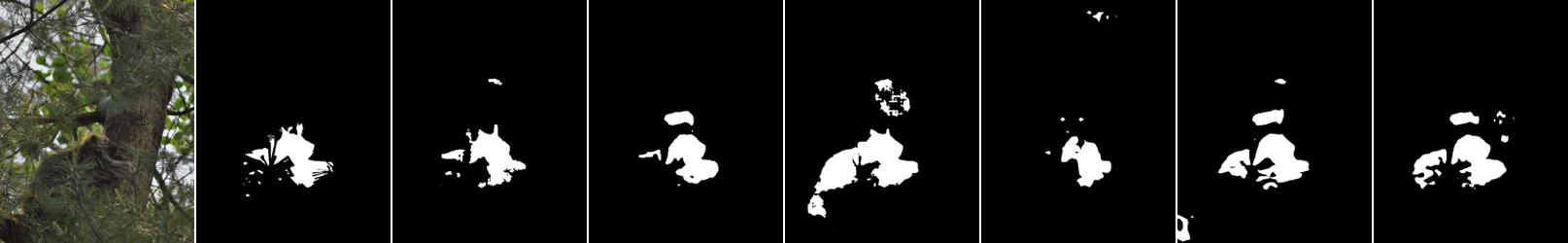}\\[0.03em]
\includegraphics[width=\textwidth]{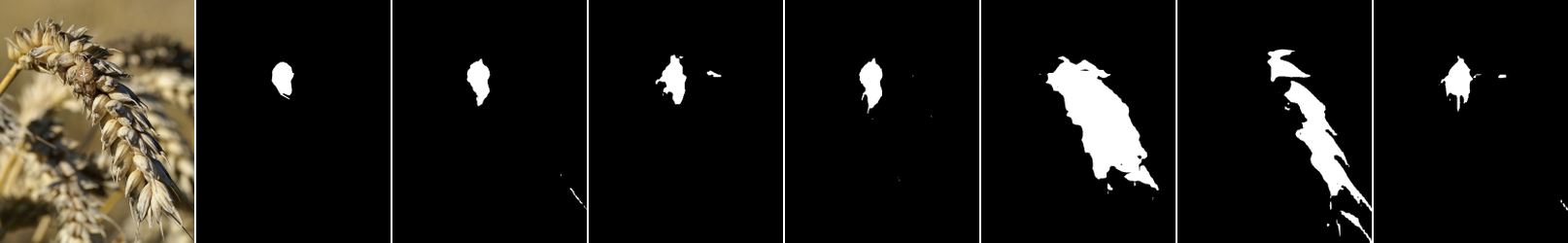}\\[0.03em]
\includegraphics[width=\textwidth]{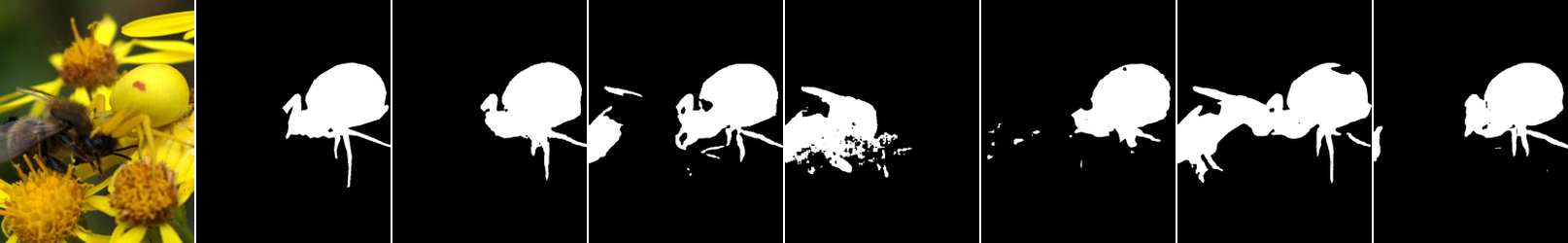}\\[0.08em]
{\small
\begin{minipage}[t]{0.125\textwidth}\centering Image\end{minipage}%
\begin{minipage}[t]{0.125\textwidth}\centering GT\end{minipage}%
\begin{minipage}[t]{0.125\textwidth}\centering Ours\end{minipage}%
\begin{minipage}[t]{0.125\textwidth}\centering ESCNet\end{minipage}%
\begin{minipage}[t]{0.125\textwidth}\centering MM-SAM\end{minipage}%
\begin{minipage}[t]{0.125\textwidth}\centering UGTR\end{minipage}%
\begin{minipage}[t]{0.125\textwidth}\centering HitNet\end{minipage}%
\begin{minipage}[t]{0.125\textwidth}\centering CamoFormer\end{minipage}%
}
\captionof{figure}{Additional qualitative comparisons with representative COD methods.}
\label{fig:supp-qual-cmp}
\end{center}

\clearpage
\begin{center}
\includegraphics[width=\textwidth]{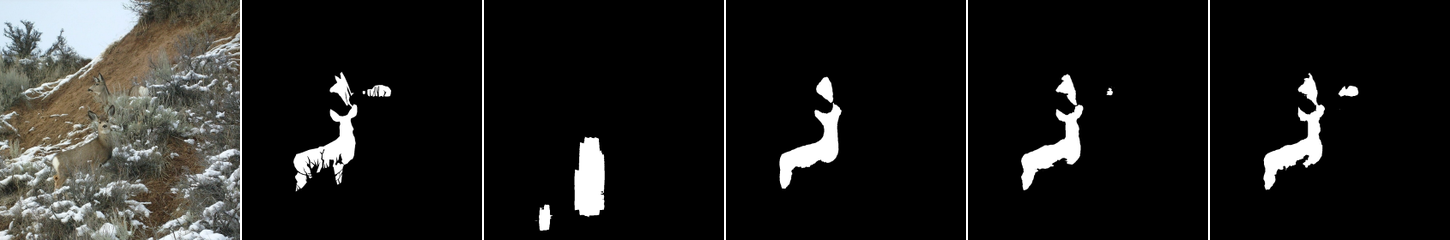}\\[0.02em]
\includegraphics[width=\textwidth]{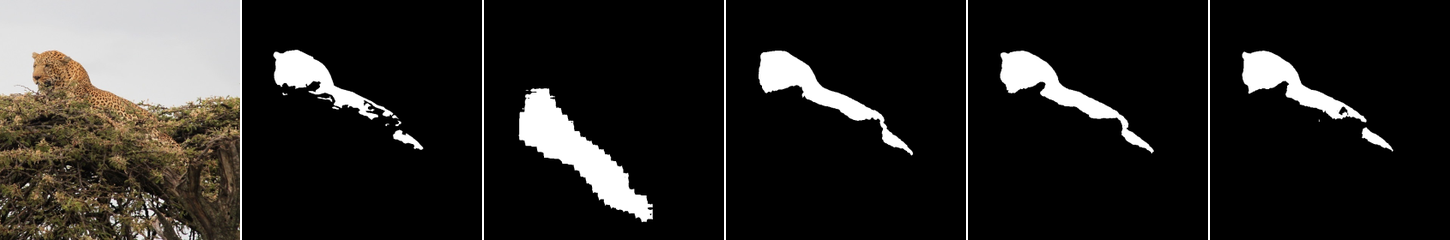}\\[0.02em]
\includegraphics[width=\textwidth]{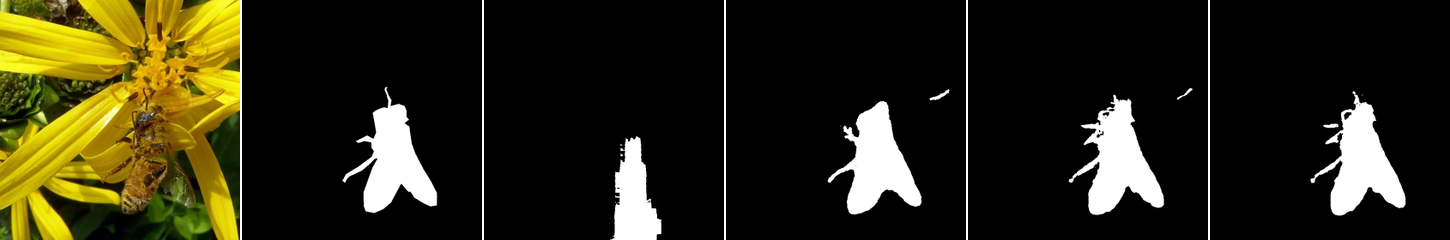}\\[0.02em]
\includegraphics[width=\textwidth]{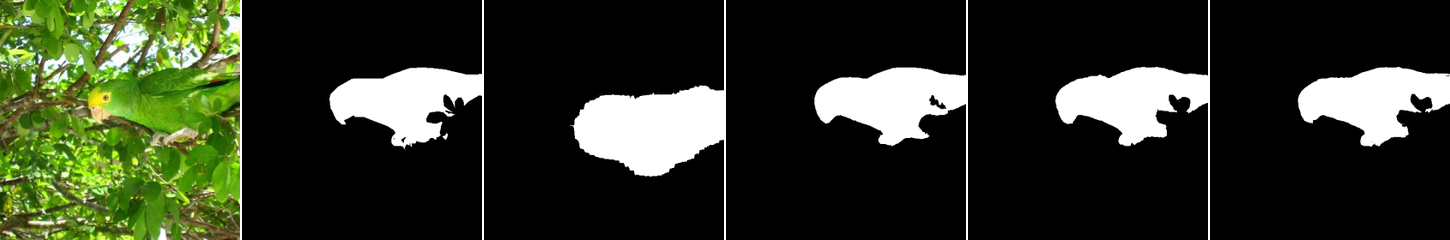}\\[0.02em]
\includegraphics[width=\textwidth]{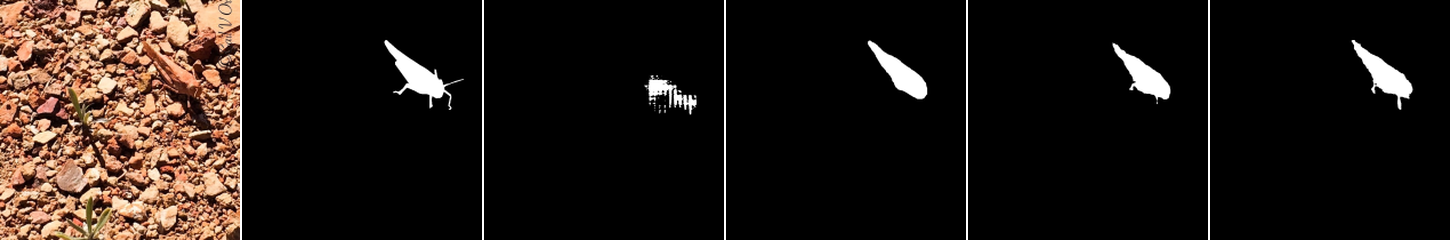}\\[0.02em]
\includegraphics[width=\textwidth]{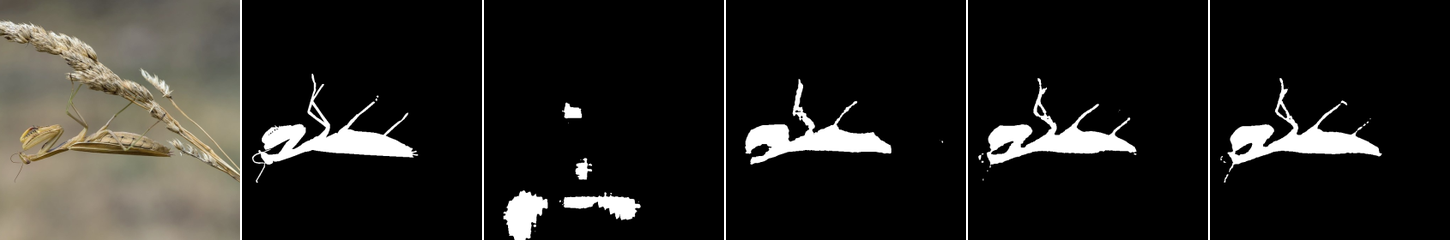}\\[0.02em]
\includegraphics[width=\textwidth]{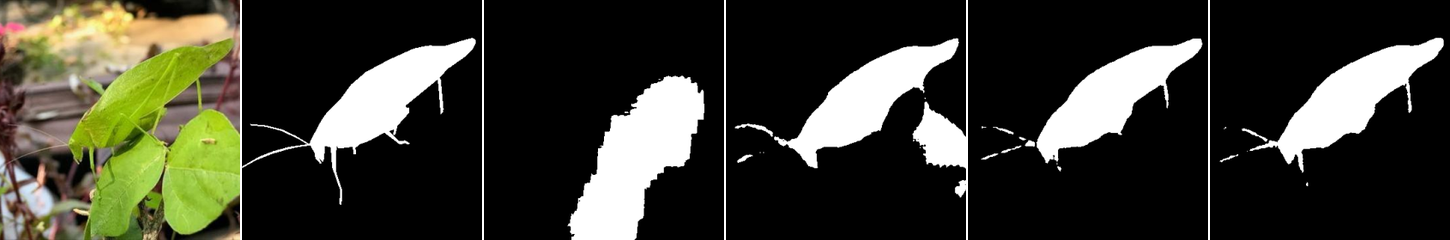}\\[0.08em]
{\small
\begin{minipage}[t]{0.166\textwidth}\centering Image\end{minipage}%
\begin{minipage}[t]{0.166\textwidth}\centering GT\end{minipage}%
\begin{minipage}[t]{0.166\textwidth}\centering Swin-SAM\end{minipage}%
\begin{minipage}[t]{0.166\textwidth}\centering $+$ target emb.\end{minipage}%
\begin{minipage}[t]{0.166\textwidth}\centering $+$ four-stage\end{minipage}%
\begin{minipage}[t]{0.166\textwidth}\centering $+$ LADVF\end{minipage}%
}
\end{center}

\clearpage
\begin{center}
\includegraphics[width=\textwidth]{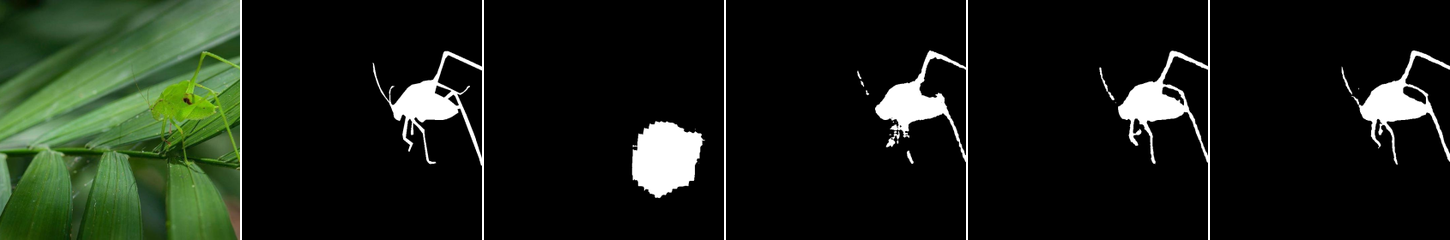}\\[0.02em]
\includegraphics[width=\textwidth]{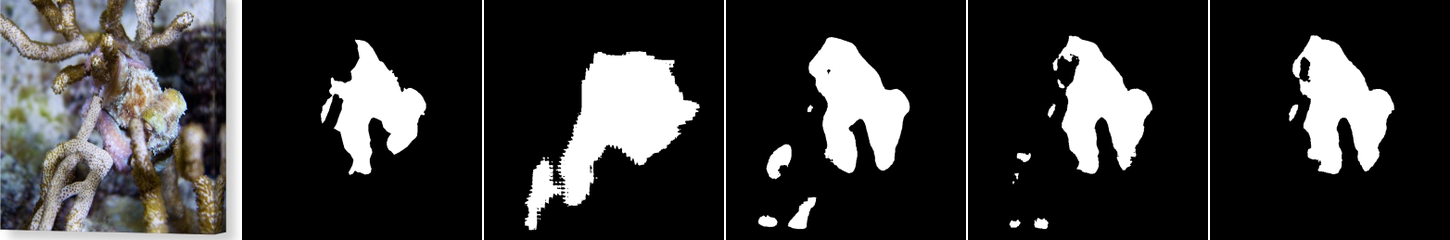}\\[0.02em]
\includegraphics[width=\textwidth]{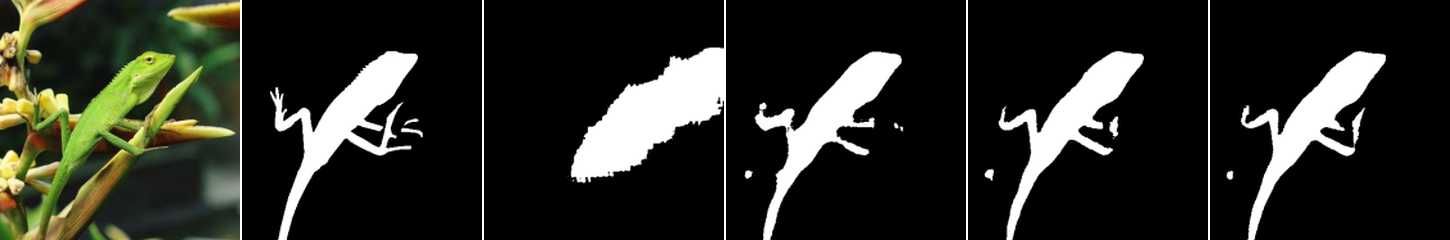}\\[0.02em]
\includegraphics[width=\textwidth]{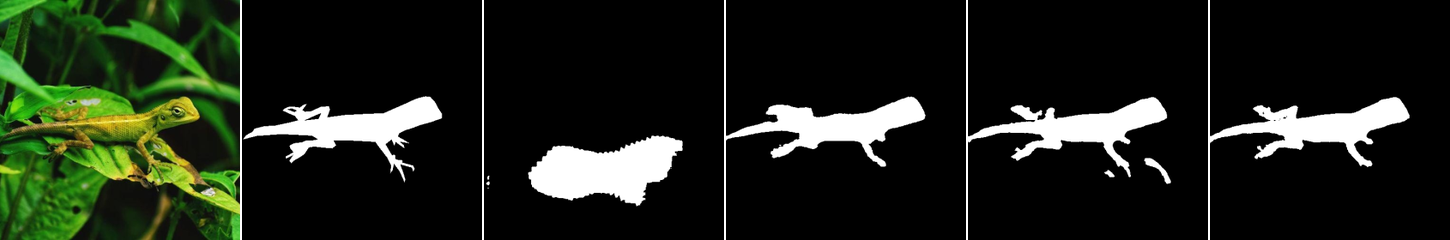}\\[0.02em]
\includegraphics[width=\textwidth]{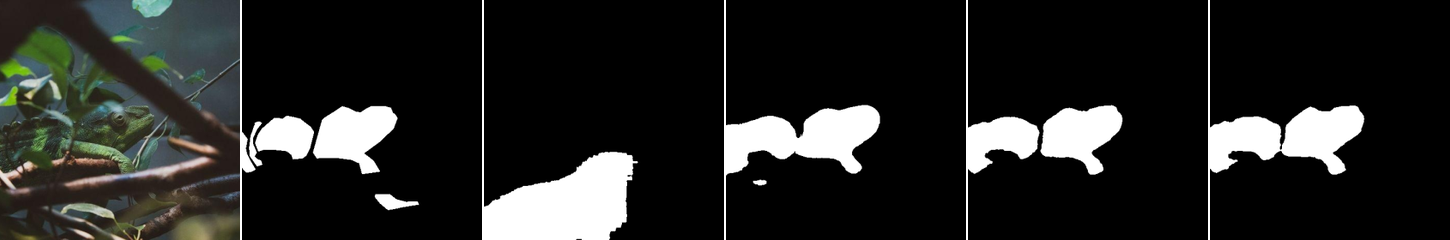}\\[0.02em]
\includegraphics[width=\textwidth]{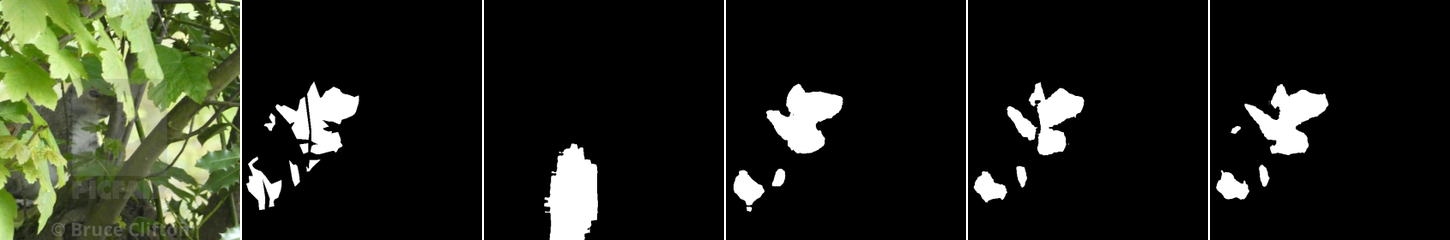}\\[0.02em]
\includegraphics[width=\textwidth]{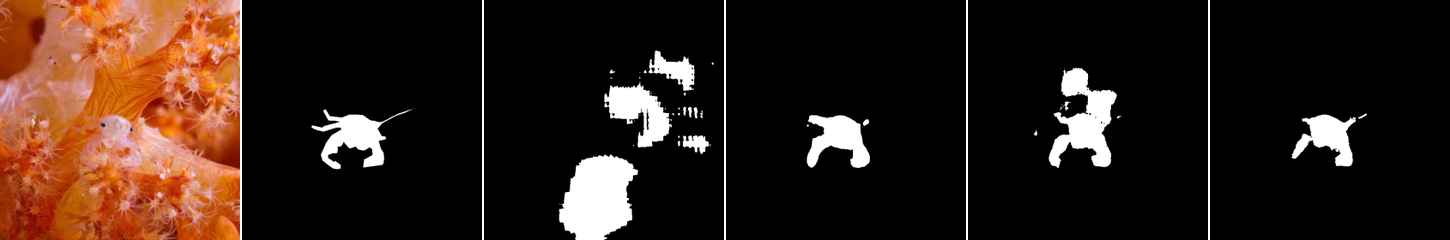}\\[0.08em]
{\small
\begin{minipage}[t]{0.166\textwidth}\centering Image\end{minipage}%
\begin{minipage}[t]{0.166\textwidth}\centering GT\end{minipage}%
\begin{minipage}[t]{0.166\textwidth}\centering Swin-SAM\end{minipage}%
\begin{minipage}[t]{0.166\textwidth}\centering $+$ target emb.\end{minipage}%
\begin{minipage}[t]{0.166\textwidth}\centering $+$ four-stage\end{minipage}%
\begin{minipage}[t]{0.166\textwidth}\centering $+$ LADVF\end{minipage}%
}
\captionof{figure}{Additional qualitative ablation results on challenging camouflaged scenes. Swin-SAM denotes the baseline that uses only the final-stage Swin feature with a SAM mask decoder (no target embedding or LADVF). Each ``$+$'' column cumulatively adds the corresponding module to the previous variant: target embedding, four-stage features, and LADVF.}
\label{fig:supp-qual-abl}
\end{center}

\clearpage
\begin{center}
\includegraphics[width=\textwidth]{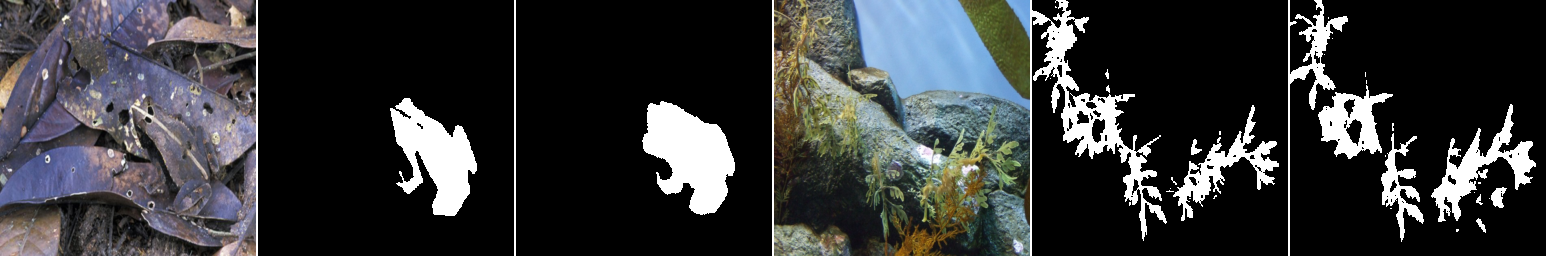}\\[0.03em]
\includegraphics[width=\textwidth]{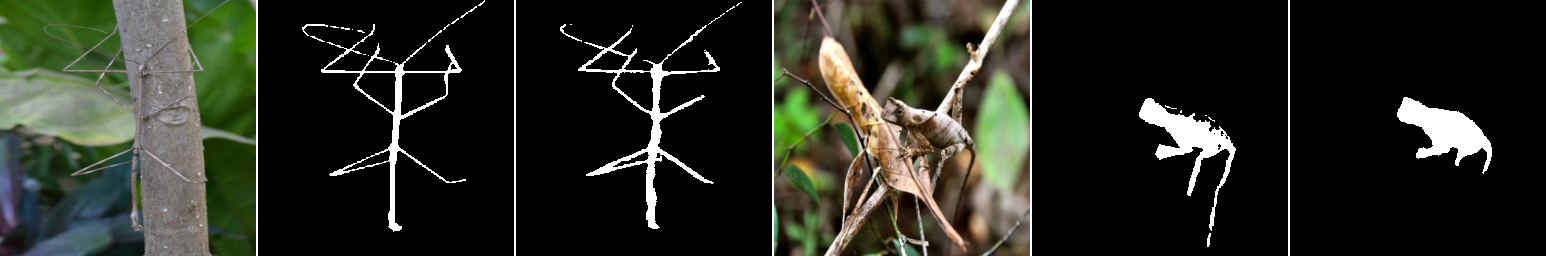}\\[0.03em]
\includegraphics[width=\textwidth]{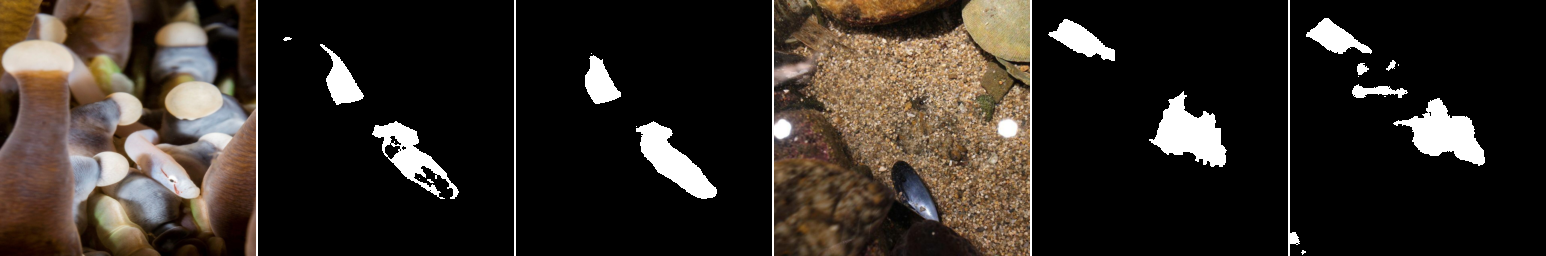}\\[0.03em]
\includegraphics[width=\textwidth]{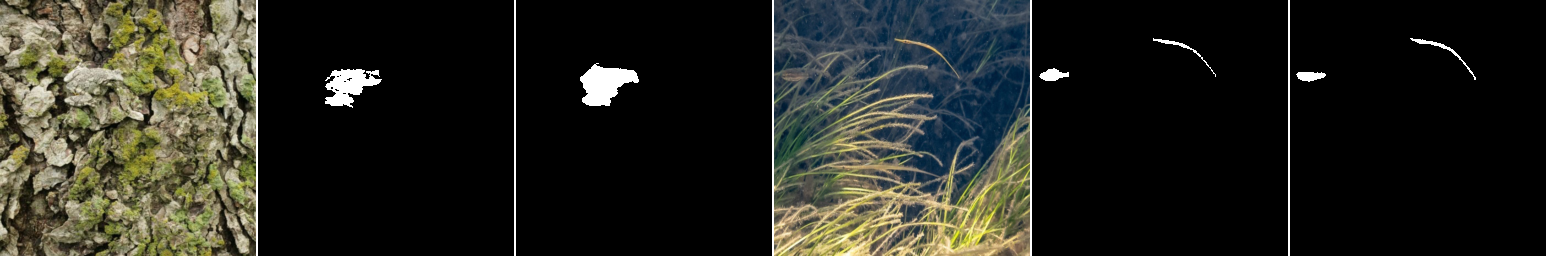}\\[0.03em]
\includegraphics[width=\textwidth]{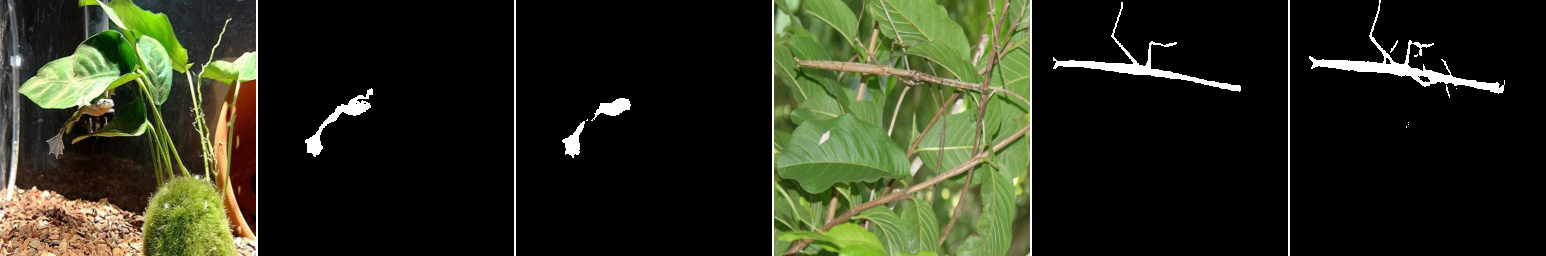}\\[0.03em]
\includegraphics[width=\textwidth]{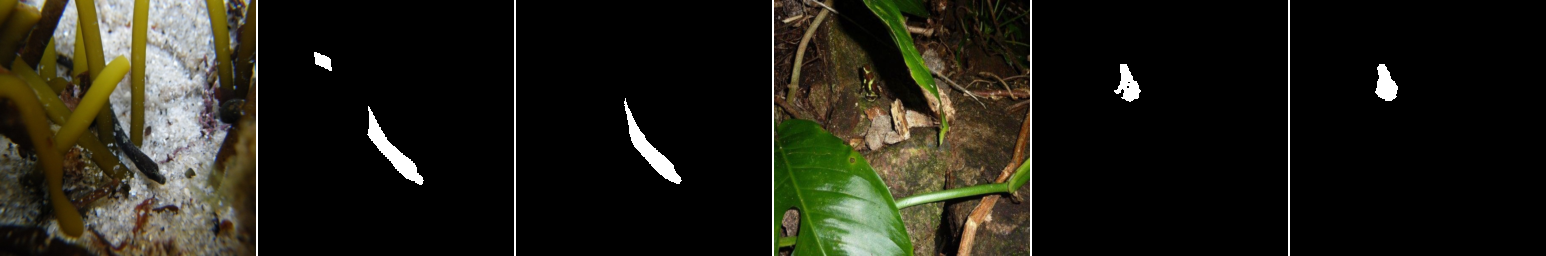}\\[0.03em]
\includegraphics[width=\textwidth]{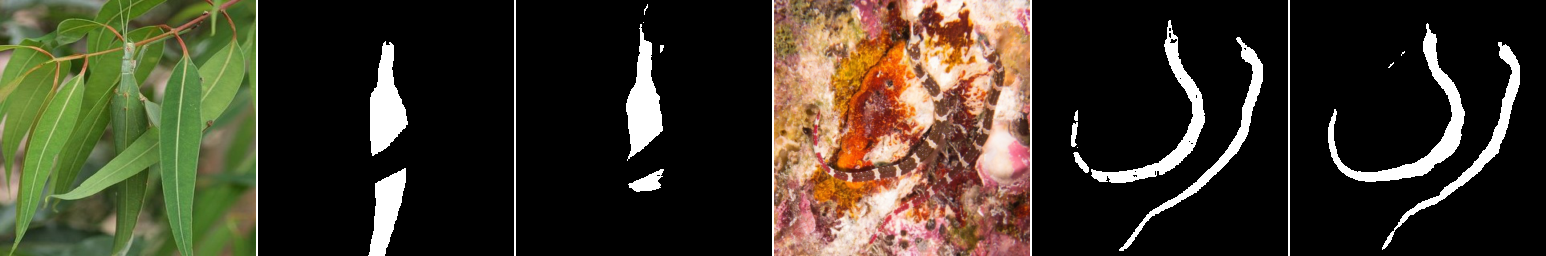}\\[0.08em]
{\footnotesize
\begin{minipage}[t]{0.166\textwidth}\centering Image\end{minipage}%
\begin{minipage}[t]{0.166\textwidth}\centering GT\end{minipage}%
\begin{minipage}[t]{0.166\textwidth}\centering Ours\end{minipage}%
\begin{minipage}[t]{0.166\textwidth}\centering Image\end{minipage}%
\begin{minipage}[t]{0.166\textwidth}\centering GT\end{minipage}%
\begin{minipage}[t]{0.166\textwidth}\centering Ours\end{minipage}%
}
\end{center}

\clearpage
\begin{center}
\includegraphics[width=\textwidth]{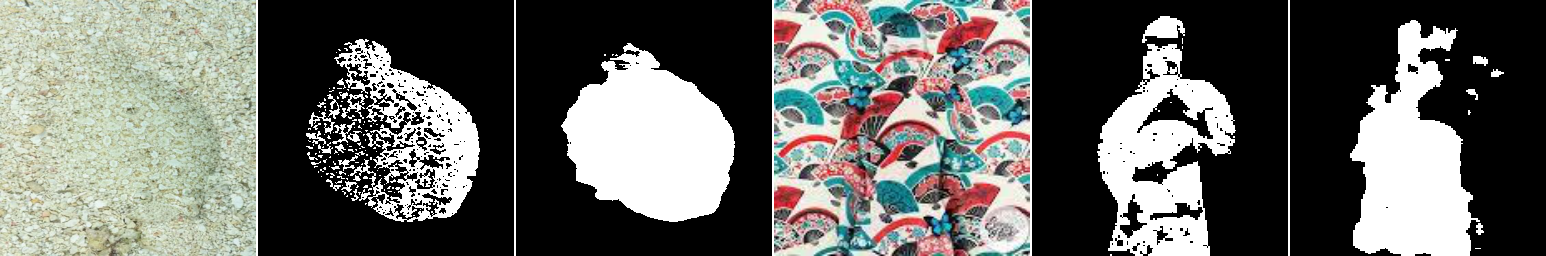}\\[0.03em]
\includegraphics[width=\textwidth]{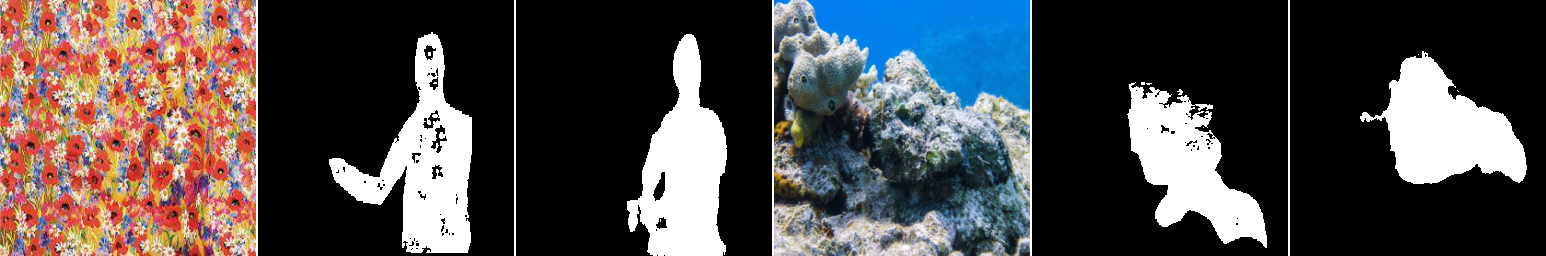}\\[0.03em]
\includegraphics[width=\textwidth]{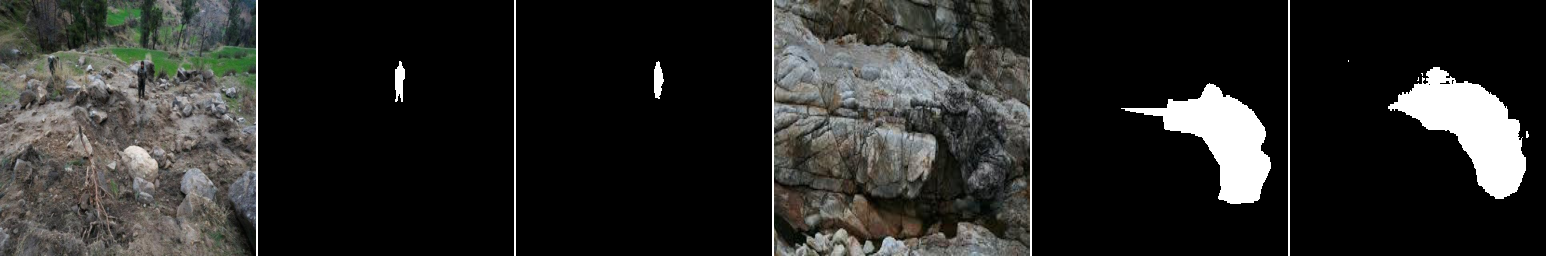}\\[0.03em]
\includegraphics[width=\textwidth]{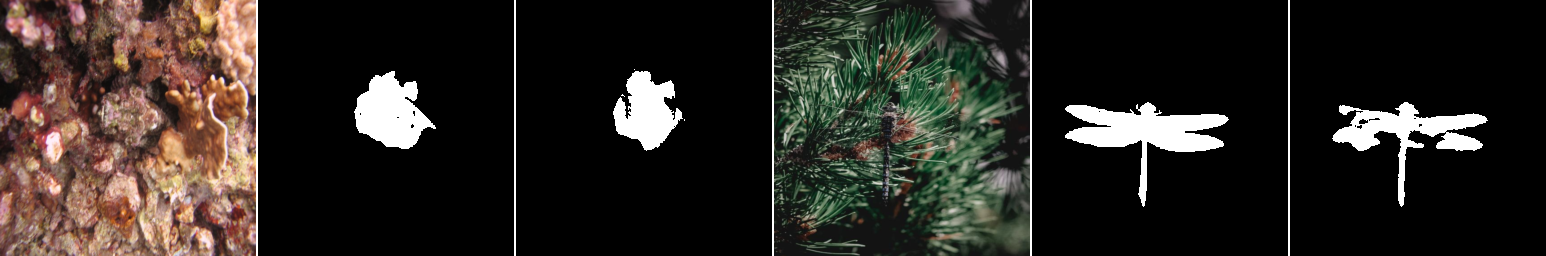}\\[0.03em]
\includegraphics[width=\textwidth]{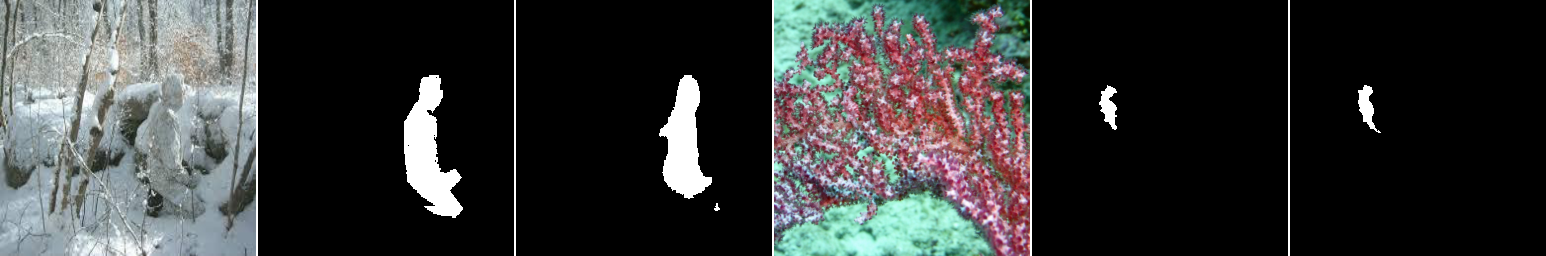}\\[0.03em]
\includegraphics[width=\textwidth]{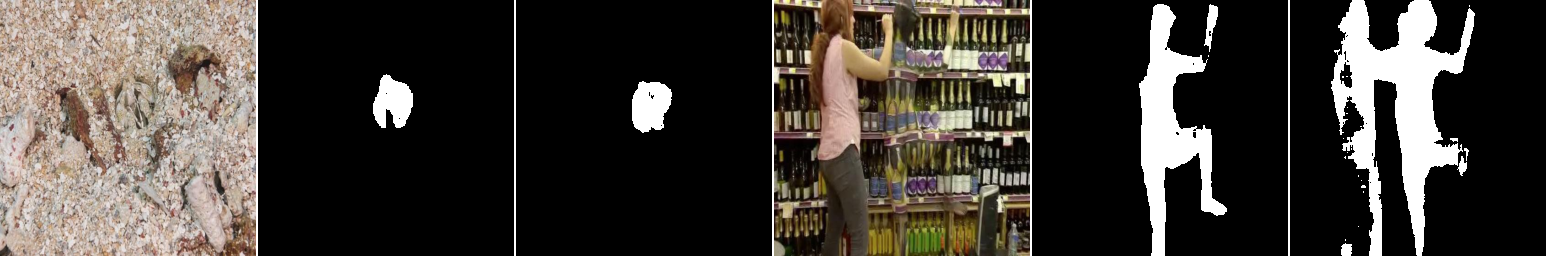}\\[0.03em]
\includegraphics[width=\textwidth]{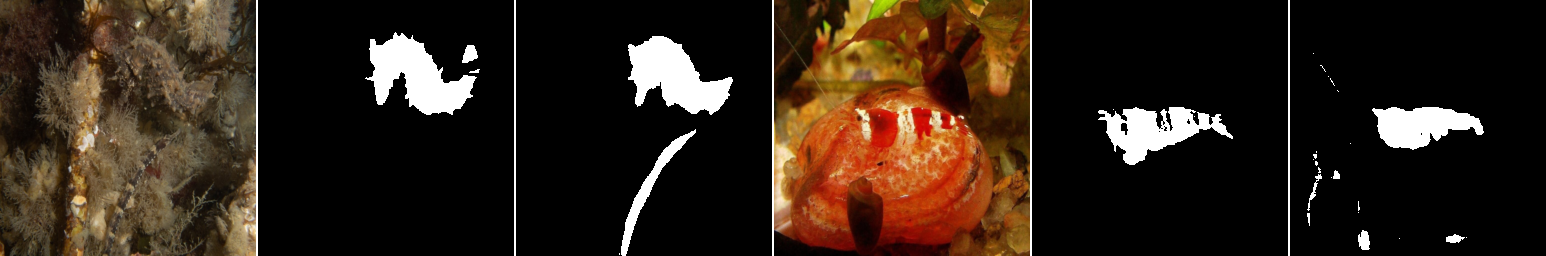}\\[0.08em]
{\footnotesize
\begin{minipage}[t]{0.166\textwidth}\centering Image\end{minipage}%
\begin{minipage}[t]{0.166\textwidth}\centering GT\end{minipage}%
\begin{minipage}[t]{0.166\textwidth}\centering Ours\end{minipage}%
\begin{minipage}[t]{0.166\textwidth}\centering Image\end{minipage}%
\begin{minipage}[t]{0.166\textwidth}\centering GT\end{minipage}%
\begin{minipage}[t]{0.166\textwidth}\centering Ours\end{minipage}%
}
\captionof{figure}{Additional Image / GT / LAD-COD mask visualizations.}
\label{fig:supp-qual}
\end{center}